\documentclass{article}

\usepackage[nonatbib]{neurips_2022}

\usepackage[utf8]{inputenc} % allow utf-8 input
\usepackage[T1]{fontenc}    % use 8-bit T1 fonts
\usepackage{hyperref}       % hyperlinks
\usepackage{url}            % simple URL typesetting
\usepackage{booktabs}       % professional-quality tables
\usepackage{amsfonts}       % blackboard math symbols
\usepackage{nicefrac}       % compact symbols for 1/2, etc.
\usepackage{microtype}      % microtypography
\usepackage{xcolor}         % colors
\usepackage{subfig}
\usepackage{graphicx}
\usepackage{graphicx}
\usepackage{amsmath}         % colors
\usepackage{csquotes}
\usepackage{tikz} 
\tikzstyle{block} = [draw, fill=yellow!40, rectangle, minimum height=3em, minimum width=4em]
\tikzstyle{sum} = [draw, fill=white, circle, scale=0.005, node distance=0.5cm]
\tikzstyle{input} = [coordinate]
\tikzstyle{output} = [coordinate]  
\tikzstyle{pinstyle} = [pin edge={to-,thin,black}]
\usetikzlibrary{shapes,arrows}
\usepackage{adjustbox}
\usepackage{multirow}
\usepackage{bbm}

\usepackage{algorithm}
\usepackage{algpseudocode}

\title{Battery GraphNets : Relational Learning for Lithium-ion Batteries(LiBs) Life Estimation}
\vspace{-3mm}
\author{
  Sakhinana Sagar Srinivas\thanks{Conceived, designed, implemented the research(programmed the software) and drafted the manuscript}, Rajat Kumar Sarkar\thanks{Performed computational experiments, interpretation, and visualization analysis of the results},\\\textbf{Venkataramana Runkana}\\
  TCS Research$^{1}$ \\
  \texttt{sagar.sakhinana@tcs.com, rajat.sarkar1@tcs.com} \\ 
  \texttt{venkat.runkana@tcs.com}
}

\begin{document}
\maketitle

\vspace{-7mm}
\begin{abstract}
\vspace{-3mm}
Battery life estimation is critical for optimizing battery performance and guaranteeing minimal degradation for better efficiency and reliability of battery-powered systems. The existing methods to predict the Remaining Useful Life(RUL) of Lithium-ion Batteries (LiBs) neglect the relational dependencies of the battery parameters to model the nonlinear degradation trajectories. We present the Battery GraphNets framework that jointly learns to incorporate a discrete dependency graph structure between battery parameters to capture the complex interactions and the graph-learning algorithm to model the intrinsic battery degradation for RUL prognosis. The proposed method outperforms several popular methods by a significant margin on publicly available battery datasets and achieves SOTA performance. We report the ablation studies to support the efficacy of our approach.
\end{abstract}

\vspace{-7mm}
\section{Introduction} 
\vspace{-3mm}
High-energy density LiBs\cite{chen2019review, manthiram2017outlook, zubi2018lithium} are driving the global revolution of battery-powered solutions 
across the whole spectrum, from portable electronics to large-scale electric mobility and energy storage systems. Recent advances in battery materials and technologies\cite{ulvestad2018brief, el2020exploits} have enabled low battery self-discharge, high power density, and lightweight, fast-charging LiBs. Despite the progress, battery capacity deteriorates\cite{o2022lithium, kim2021atom, zhou2021lithium, mohtat2021algorithmic} due to repetitive charging and discharging, leading to low performance, 
low storage characteristics, and shorter life of the LiBs. The development of accurate battery RUL\cite{xiong2018towards, hasib2021comprehensive, wu2016online} estimation algorithms has gained attention in the advancement of battery management systems(BMS, \cite{liu2019brief, shen2019review}) as the RUL translates to reliability, available range, and performance of battery-powered systems. The existing battery RUL prediction methods\cite{saha2009modeling, daigle2016end, nuhic2018battery, andre2013comparative} are limited in modeling the non-linear behavior of discharge characteristics of aging LiBs for accurate RUL prognosis.  We begin with the hypothesis that LiBs are stochastic dynamical systems. The complex interactions between multiple degradation mechanisms, characterized by different physical and chemical processes, are responsible for battery degradation. The collective behavior of the interconnected and individual degradation mechanisms, as a whole, give rise to the non-linear dynamics of the complex system. The battery parameters capture the measurable effects of tractable battery degradation. There is a need and necessity to learn the system's elusive, relational structural-dynamic dependencies of the interdependent battery parameters while simultaneously learning the non-linear degradation dynamics of LiBs. The literature on relational learning or reasoning tasks is overwhelmingly focused on dynamical learning of interacting systems dynamics using the following approaches. (a) The implicit interaction models\cite{sukhbaatar2016learning, guttenberg2016permutation, santoro2017simple, watters2017visual, hoshen2017vain} utilize a fully-connected graph to model the relational structure. Redundant edges and learning noise-incorporated representations are the drawbacks of implicit interaction models.  (b) The explicit interaction models are categorized into the following methods. (1) continuous relaxation methods\cite{jiang2019semi, wang2020gcn, chen2020iterative, zhang2020gnnguard} that utilize graph similarity metric learning to obtain the real-valued weighted adjacency matrix, and (2) probabilistic sampling approaches\cite{franceschi2019learning, zheng2020robust, kazi2022differentiable} model the edge probability distribution over graphs. 
The relaxation methods prunes edges according to edge weights and implicitly impose sparse constraint. The probabilistic methods sample to obtain a discrete graph structure by adopting reparameterization tricks\cite{bengio2013estimating, kool2019stochastic, jang2016categorical}. These methods suffer from poor convergence speed, sensitivity to the hyperparameters, etc. In this work, we present the Battery GraphNets framework(BGN) for the RUL prognosis of LiBs and operate in a bi-level approach. First, the framework utilizes graph as a mathematical model to learn the underlying pairwise relations of the interacting battery parameters in the observed data by optimizing the differentiable discrete graph structure of battery parameters. It then optimizes the trainable parameters of the graph learning algorithm via the back-propagation algorithm with a gradient descent technique that operates on the inferred relational graph topology to model the time-evolving dynamics of LiBs for improved RUL estimates. The point-wise forecasting error of RUL prediction acts as deep-supervision information of the underlying  graph structures. Intuitively, BGN learns the fixed-size node representations by propagating and aggregating neural messages between neighboring nodes according to the graph structure, wherein the interactions are captured implicitly in the neural messages. In short, the unified prognostic framework simultaneously learns the latent graph structure of multiple interacting battery parameters and the Graph Neural Networks(GNNs) backbone parameters. GNNs model the observations of the dynamical interacting system to facilitate the downstream RUL prognosis with predictive uncertainty estimation of LiBs. 
 
\vspace{-6mm} 
\section{Problem statement} 
\vspace{-5mm}
Consider an arbitrary training set of graph-structured battery data consisting of graph-label pairs $\{(\mathcal{G}^{t}, y^{t})\}$. The goal is to learn a model($f:\mathcal{G}^{t} \rightarrow  y^{t}$) trained in inductive supervised learning to predict the label $y^{t}$ at any time step $t$ associated with the input graph $\mathcal{G}^{t}$ and estimate uncertainties. In short, the input to the model is the multivariate battery parameters measurements viewed as the sequence of graphs $\mathcal{G}^{t}$ and predicts the RUL estimates $y^{t}_{p}$. 

\vspace{-6mm} 
\section{Proposed Approach}  
\begin{figure}[htbp]
\centering
\hspace*{-32.5mm}\includegraphics[width=1.40\textwidth]{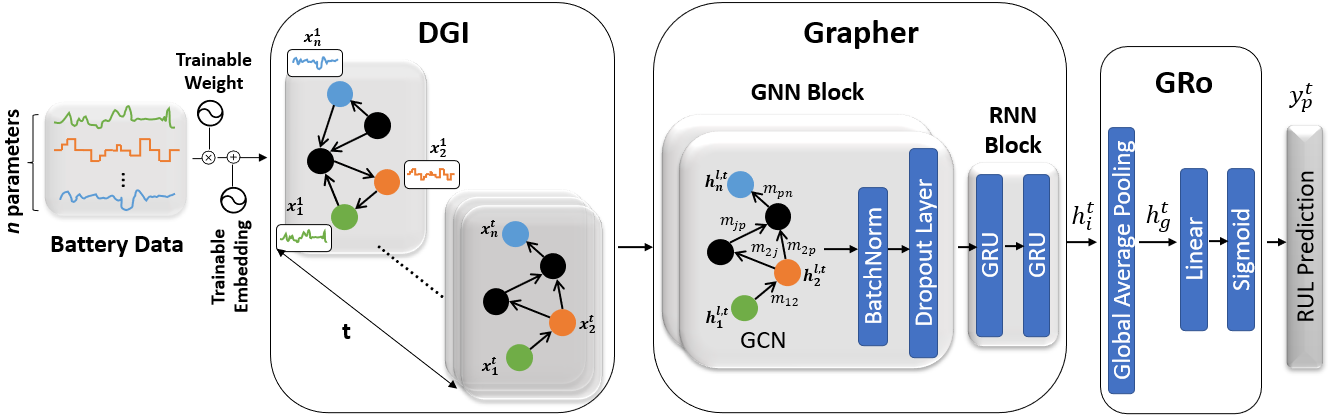}
\vspace{-5mm}
\caption{The Battery GraphNets(BGN) framework. (a) The DGI module learns the dynamic graphs. (b) The Grapher module consists of GNN and RNN blocks. The GNN block operates on the dynamic graph topology to encode the local-graph neighborhood information in the node-level representations. The RNN block regulates the information flow to capture long-range spatio-temporal dependencies in the node representations. (c) The Graph Readout(GRo) module computes the entire graph representation to preserve the global graph information. $y^{t}_{p}$ denotes the model predictions.}
\label{fig:a}
\end{figure}

\vspace{-5mm}
The proposed framework consists of the following modules. (a) The dynamic graph inference module, for brevity, DGI. It is a probabilistic model that learns the time-evolving optimal graph structure for the structured spatio-temporal representation of the battery data. (b) The grapher module presents a node-level graph encoder that computes the fixed-size representation for each node by jointly modeling the latent inter-parameters and the long-range, intra-parameters dependencies in the graph-structured battery data. (c) A graph readout module for global pooling of the node-level representations to capture the global semantic structure in the fixed-length graph-level vector. We utilize a linear projection to transform the entire-graph fixed-size vector and apply a non-linear activation function to predict the high-quality RUL estimates. In summary, the proposed framework simultaneously learns the spatio-temporal dependencies among multiple battery parameters over time with a GNN backbone operating on a discrete latent graph in the node-level representations. It then performs inference over the node-level latent variables to predict RUL. Figure \ref{fig:a} illustrates the proposed method.

\vspace{-4mm}

\vspace{-7mm}
\subsection{Dynamic Graph Inference(DGI)} 
\vspace{-2mm} % 
The battery parameters do not have an explicit relational graph structure(unknown or partially known) underlying the data. We define a dynamic graph, $\mathcal{G}^{t}$ at a time point, $t$ as a tuple($\mathcal{V}, \mathcal{E}^{t}, \{\mathbf{x}^{t}_{i}\}_{i \in \mathcal{V}}$). $\mathcal{V}$ denotes a set of nodes. We represent each battery parameter as a node $i \in \mathcal{V}$.  Every node is associated with a feature vector $\mathbf{x}^{t}_{i} \in \mathbb{R}^{\mathcal{C} \cdot \mathcal{T}}$. Here, $\mathcal{C}$ denotes the number of charge-discharge cycles of the LiBs, and $\mathcal{T}$ is the number of time steps in each charge cycle. $\mathcal{E}^{t}$ is the set of (directed) edges where each edge $e_{ji} \in \mathcal{E}^{t}$ models the pairwise relationship between the neighboring nodes $j,i$. The dynamic graphs $\mathcal{G}^{t}$ have a feature matrix, $\mathbf{X}^{t}$ that changes dynamically over time, an adjacency matrix, $\mathcal{A}^{t}$, and an edge-set, $\mathcal{E}^{t}$ that may vary over time, whereas the node-set $\mathcal{V}$ remains unchanged. The nodes of the dynamic graph are associated with trainable embeddings $\mathbf{b}_i \in \mathbb{R}^{d}$, $1 \leq i \leq n$, which are $d$-dimensional continuous vector representations. DGI module learns the sparse regularized dynamic-graph structure from the explicit-battery data, $\mathbf{x}^{t}_{i}$ integrated with the implicit node embeddings, $\mathbf{b}_i$. We compute the pairwise parameters as given by,

\vspace{-4mm}
\resizebox{1\linewidth}{!}{
\begin{minipage}{\linewidth}
\begin{align}
\hspace{20mm}\theta^{t,k}_{i,j} &= \sigma\bigg(g_{fc} \bigg(\big(\mathbf{b}_i + \mathbf{W}_{s}\mathbf{x}^{t}_{i} \big) || \big(\mathbf{b}_j + \mathbf{W}_{s}\mathbf{x}^{t}_{j}\big)\bigg) \bigg) , \forall i,j \in\{1, \ldots, n\}, k \in\{0,1\} \nonumber \label{eq:DGI}
\end{align}
\end{minipage}
}

\vspace{-2mm}
Where $\theta^{t,k}_{i,j} \in \mathbb{R}^{n^{2} \times 2}$ denotes probability matrix, $\mathbf{W}^{d \times (\mathcal{C} \cdot \mathcal{T})}_{s}$ is trainable weight matrix. $g_{fc}$ is a stack of fully-connected layers. $\sigma$ is a logistic sigmoid function. $n$, i.e., $|\mathcal{V}|$ denotes number of battery parameters. $||$ denotes vector concatenation. $\theta^{t,k}_{i,j}$ encodes relation between a pair of nodes $\left(i, j\right)$ to a scalar $\in[0,1]$. $\theta^{t,0}_{i,j}$ represents probability of a directed edge from node $i$ to its potential neighbor $j$ at a time point $t$, and $\theta^{t,1}_{i,j}$ represents contrariwise probability. We sample the time-specific adjacency matrix\cite{shang2021discrete}, i.e., $\mathcal{A}^{t}_{i,j} \in \mathbb{R}^{n \times n}$ to determine the edges to preserve and drop the redundant edges via the element-wise, differentiable method, using the Gumbel-softmax trick\cite{jang2016categorical, maddison2016concrete} as described by,

\vspace{-3mm}
\resizebox{1\linewidth}{!}{
\begin{minipage}{\linewidth}
\begin{align}
\mathcal{A}^{t}_{i,j} =\exp \bigg(\big(g_{i,j}^k + \theta^{t,k}_{i,j}\big) / \gamma\bigg)\bigg/{\sum_k \exp \bigg(\big(g_{i,j}^k+ \theta^{t,k}_{i,j}\big) / \gamma\bigg)}
\end{align}
\end{minipage}
}

\vspace{-2mm}
where $g_{i j}^k \sim \operatorname{Gumbel}(0,1) = \log (-\log (U(0,1))$ where $U$ is uniform distribution. $\gamma$, denotes the temperature parameter with 0.05. In summary, the DGI module learns the evolving irregular graph structure, which captures the time-dependent interdependence relations of the battery parameters. By viewing the battery data as node-attributed discrete-time dynamic graphs, we explore the GNN backbone to encode the complex discrete graph-structured data by summarizing the entire graph's structural characteristics to compute its optimal representations to facilitate the RUL prediction task.

\vspace{-5mm}
\subsection{Grapher} 
\vspace{-3mm}   
The grapher module consists of two blocks (1) the GNN block and the (2) the RNN block. The GNN block consists of a single-layer GNN $\rightarrow$ Batch Norm $\rightarrow$ Dropout sequence. We apply batchnorm\cite{ioffe2015batch} and dropout\cite{baldi2013understanding} for regularization. The GNN layer-wise operator involves two basic node-level computations at every layer. (1) AGGREGATE operation: aggregates neural messages($\mathbf{m}_{ji}$) from local-graph neighbors $j \in \mathcal{N}_i$, where $\mathbf{m}_{ji}$ denotes the neural message dispatched from source node $j$ to sink node $i$ via edge $e_{ji}$. It captures the lateral interaction. (2) UPDATE operation:  transforms each node representation by combining its fixed-length representation obtained in the previous layer and the aggregated neural messages. We utilize the vanilla variant of GCN\cite{kipf2016semi} to model the GNN layer, which couples the aggregate and update operations to refine the node-level representations. Let's consider an $\text{L}$-layer GCN, the mathematical formulation of the $l^{th}$ layer ($1 \leq l \leq \text{L}$) described below,

\vspace{-7mm}
\resizebox{1\linewidth}{!}{
\begin{minipage}{\linewidth}
\begin{align}
\mathbf{h}_i^{(l, t)}=\operatorname{ReLU}\bigg(\mathbf{W}^{(l)}_{g} \hspace{-1mm} \cdot \hspace{-2mm} \sum_{j \in \mathcal{N}(i) \cup\{i\}} \hspace{-1mm}\frac{1}{\sqrt{d^{t}_i d^{t}_j}} \mathbf{h}_j^{(l-1, t)}\bigg)
\end{align}
\end{minipage}
}

\vspace{-3mm}
where $\mathbf{h}_i^{(l, t)} \in  \mathbb{R}^{d}, i \in \mathcal{V}$ denotes node representation in the $l^{th}$ layer with $\mathbf{h}_i^{(0, t)}= \mathbf{W}_{s}\mathbf{x}^{t}_i$. $\mathbf{W}_{g} \in  \mathbb{R}^{d \times d}$ is weight matrix. $d_{i}$ represents the degree of node $i$. We stack two GNN blocks and concatenate the output node-level representations from each block to compute $\mathbf{h}_i \in \mathbb{R}^{2d}$. The RNN block consists of double-stacked GRU units\cite{cho2014learning, chung2014empirical} that operate on  $\mathbf{h}_i \in \mathbb{R}^{2d}$ for transforming the node-level representations to compute $\mathbf{h}_i \in \mathbb{R}^{d}$. It regulates the information flow through the gating mechanism to better capture the long-range spatio-temporal dependencies. 

\vspace{-5mm}
\subsection{Graph Readout(GRo)}
\vspace{-3mm} 
We utilize the global average pooling operation to aggregate node-level representations($\mathbf{h}^{t}_i$) computed by the grapher module and obtain a graph-level representation $\mathbf{h}_g$, as given by

\vspace{-4mm}
\resizebox{1\linewidth}{!}{
\begin{minipage}{\linewidth}
\begin{align}
\mathbf{h}^{t}_g=\operatorname{READOUT}\bigg(\big\{ \big(\mathbf{h}_i^{t} \odot \mathbf{b}_i \big) \mid i \in \mathcal{V}\big\}\bigg)  \label{eq:GRo}
\end{align}
\end{minipage}
}

\vspace{-1mm}
where $\odot$ denotes element-wise multiplication. We utilize a linear layer and apply a sigmoid activation function to transform $\mathbf{h}^{t}_g$ and predict the  RUL($y^{t}_{p}$) at a time point, $t$. 

\vspace{-3mm}
\subsection{Uncertainty Estimation}
\vspace{-2mm}
The BGN framework is trained in a supervised learning approach to minimize the mean squared error deviations between model RUL estimates $y^{t}_{p}$ and ground-truth $y^{t}$ as described below,

\vspace{-4mm}
\begin{equation}
\mathcal{L}_{\text{MSE}} = \frac{1}{\beta}\sum_{t=1}^{\beta}(y^{t}_{p}-y^{t})^2 \label{eq:UCE}
\end{equation}

\vspace{-2mm}
where $\beta = \mathcal{C}_{train} \cdot \mathcal{T}_{train}$ denotes total time steps in the training set. We present the BGN framework with predictive uncertainty estimation for short, BGN-UE that computes the mean($\mu_\phi\left(\mathbf{x}_i^{t}\right)$) and the variance($\sigma_\phi^2\left(\mathbf{x}_i^{t}\right)$) of the Gaussian distribution and regards RUL estimate as a sample from this distribution. We predict the outputs  of the BGN-UE framework, $\mu_\phi\left(\mathbf{x}_i^{t}\right), \sigma_\phi^2\left(\mathbf{x}_i^{t}\right)$ as given by,

\vspace{-4mm}
\resizebox{1\linewidth}{!}{
\begin{minipage}{\linewidth}
\begin{align}
\mu_\phi\left(\mathbf{x}_i^{t}\right), \sigma_\phi^2\left(\mathbf{x}_i^{t}\right) &= f_\theta\bigg(\big\{ \big(\mathbf{h}_i^{t} \odot \mathbf{b}_i \big) \mid i \in \mathcal{V}\big\}\bigg) 
\end{align}
\end{minipage}
}

\vspace{0mm}
where $\phi$ represents trainable parameters of the BGN-UE framework and $f_{\theta}$ is modeled by the stack of fully-connected layers.  The  RUL prediction, $\hat{y}^{t}_{p} \sim \mathcal{N}\big(\mu_\phi\left(\mathbf{x}_i^{t}\right), \sigma_\phi^2\left(\mathbf{x}_i^{t}\right)\big)$ at the $t^{th}$ time point is the maximum likelihood estimate(MLE) of the predicted Gaussian distribution, i.e., its mean at time point $t$, i.e., $\hat{y}^{t}_{p} = \mu_\phi\left(\mathbf{x}_i^{t}\right)$. The BGN-UE framework optimizes the negative Gaussian log likelihood\cite{nix1994estimating} of the battery parameters observations based on its predicted mean $\big(\mu_\phi\big(\mathbf{x}^{t}_i\big)\big)$ and variance $\big(\sigma_\phi^2\big(\mathbf{x}^{t}_i\big)\big)$ as described by,

\vspace{-4mm}
\resizebox{1.0\linewidth}{!}{
\begin{minipage}{\linewidth}
\begin{equation}
\mathcal{N}(y^{t};\mu_\phi\big(\mathbf{x}^{t}_i\big),\sigma_\phi\big(\mathbf{x}^{t}_i\big))={\frac {1}{\sigma_\phi\big(\mathbf{x}^{t}_i\big) {\sqrt {2\pi }}}}\hspace{1.5mm} e^{-{\dfrac {1}{2}}\left({\dfrac {y^{t}-\mu_\phi\big(\mathbf{x}^{t}_i\big) }{\sigma_\phi\big(\mathbf{x}^{t}_i\big)}}\right)^{2}}
\end{equation} 
\end{minipage}
}

We apply logarithm transformation on both sides of the equation, 

\vspace{-5mm}
\resizebox{1.0\linewidth}{!}{
\begin{minipage}{\linewidth}
\begin{align}
\log\ \mathcal{N}(y^{t};\mu_\phi\big(\mathbf{x}^{t}_i\big),\sigma_\phi\big(\mathbf{x}^{t}_i\big)) &= \log\left[{\frac {1}{\sigma_\phi\big(\mathbf{x}^{t}_i\big) {\sqrt {2\pi }}}}\right] + \log \left[e^{-{\dfrac {1}{2}}\left({\dfrac {y^{t}-\mu_\phi\big(\mathbf{x}^{t}_i\big) }{\sigma_\phi\big(\mathbf{x}^{t}_i\big)}}\right)^{2}}\right] \\
 &= \log\ {\frac {1}{\sigma_\phi\big(\mathbf{x}^{t}_i\big)}} + \log\ {\frac {1}{{\sqrt {2\pi }}}} -{\frac {1}{2}}\left(\dfrac {y^{t}-\mu_\phi\big(\mathbf{x}^{t}_i\big) }{\sigma_\phi\big(\mathbf{x}^{t}_i\big)}\right)^{2} \\
 &= -\log\ \sigma_\phi\big(\mathbf{x}^{t}_i\big) + C -{\frac {1}{2}}\left(\dfrac {y^{t}-\mu_\phi\big(\mathbf{x}^{t}_i\big) }{\sigma_\phi\big(\mathbf{x}^{t}_i\big)}\right)^{2}
\end{align}
\end{minipage}
}

We drop the constant(C) and the Gaussian negative log likelihood loss(i.e., negative log gaussian probability density function(pdf)) for the full training dataset is described by, 

\vspace{-5mm}
\resizebox{1.05\linewidth}{!}{
\begin{minipage}{\linewidth}
\begin{align}
\mathcal{L}_{\text{GaussianNLLLoss}} = \frac{1}{2} \sum_{t=1}^{\beta}\left[\log\ \sigma_\phi\big(\mathbf{x}^{t}_i\big)^2 + {\frac {(y^{t}-\mu_\phi\left(x^{t}_i\right))^{2}}{\sigma_\phi\left(x^{t}_i\right)^2}}\right] \nonumber
\end{align}
\end{minipage}
}

\vspace{-3mm}
\resizebox{1\linewidth}{!}{
\begin{minipage}{\linewidth}
\begin{align}
\hspace{10mm}\mathcal{L}_{\text{GaussianNLLLoss}} =  \sum_{t=1}^\beta \left[\frac{\log \sigma_\phi\big(x^{t}_i\big)^2}{2}+\frac{\left(y^{t}-\mu_\phi\left(x^{t}_i\right)\right)^2}{2 \sigma_\phi\left(x^{t}_i\right)^2}\right] \label{eq:GUCE}
\end{align}
\end{minipage}
}

\vspace{0mm}
Note: The neural network($f_{\theta}$) predicts the mean($\mu_\phi\left(\mathbf{x}_i^{t}\right)$) and standard deviation($\sigma_\phi^2\left(\mathbf{x}_i^{t}\right)$) of a normal distribution modeling the target RUL, $y^{t}$. In summary, the BGN framework minimizes $\mathcal{L}_{\text{MSE}}$(refer Equation \ref{eq:UCE}) and the BGN-UE framework minimizes $\mathcal{L}_{\text{GaussianNLLLoss}}$(refer Equation \ref{eq:GUCE}) respectively.

\vspace{-4mm} 
\section{Experiments and Results}
\label{expresults}
\vspace{-3mm}
\subsection{Datasets \& Experimental Settings}
\vspace{-2mm}
We validate the proposed frameworks on two publicly available datasets: the NASA Randomized Battery Usage Dataset\cite{saha2007battery} and the UNIBO Powertools Dataset\cite{wong2021li}. We utilize the predefined train/test datasets provided by the dataset curators for competitive benchmarking with the varied baseline methods. The battery datasets consist of the following parameters: voltage, current, charge capacity, discharge capacity, charge energy and discharge energy. The training dataset consists of two separate sets, (1) the independent features $\in \mathbb{R}^{n \times \beta}$, and (2) the target to predict $\in \mathbb{R}^{\beta}$. The target to predict is the predetermined RUL, the set of ground-truth labels corresponding to the independent features of the training dataset. We utilize the data preprocessing implementation reported in \cite{michael2022battery, bosello2021charge}. We performed the computational experiments on the preprocessed datasets. We leveraged the k-fold cross-validation technique with k = 10 to learn the optimum model parameters on the training/validation dataset. We implement the early stopping technique on the validation dataset to tune hyperparameters and prevent overfitting. The model was trained for 100 epochs with a batch size of 48 on multiple NVIDIA Tesla GPUs to learn from the training dataset. The initial learning rate($\textit{lr}$) was $1e^{-3}$. The embedding size(d) was 32. We leveraged the learning rate scheduler to drop the $\textit{lr}$ by half if the model performance stops improving for a patient number of 10 epochs on the validation dataset. We utilize the Adam optimization algorithm\cite{kingma2014adam} to minimize the (a) mean-squared error loss for the BGN model and (b) the negative Gaussian log-likelihood for the BGN-UE model. We evaluated the model performance and report the results on the test dataset. We conducted five independent experimental runs and report the ensemble average of the results.

\vspace{-4mm}
\subsection{Benchmarking algorithms}%
\vspace{-3mm}
Table \ref{tab:baseline} reports the performance of the state-of-the-art methods compared to our proposed models on both datasets. The performance evaluation metric is RMSE. The list of baseline models includes several graph-learning techniques, Graph Neural Networks(GNNs, \cite{Fey/Lenssen/2019}), Graph Temporal Algorithms(GTA, \cite{rozemberczki2021pytorch}), and sequence models such as Recurrent Neural Networks(RNN, \cite{hochreiter1997long, cho2014properties, chung2014empirical}), Transformer networks\cite{vaswani2017attention}, etc. The results show that our proposed models attains significant gains w.r.t to the lower value of the evaluation metric on all datasets. The experimental results support the rationale of our framework to learn the time-varying dependencies and dynamics of the LiBs with a GNN backbone to provide better RUL estimates compared to the baseline methods. We brief the baseline algorithms and report additional results in the appendix.

\vspace{-5mm}
\begin{table}[!htbp]
\centering
\caption{Comparative study of our proposed method and the baseline algorithms.}
\label{tab:baseline}
\setlength{\tabcolsep}{4pt}
\resizebox{1.035\textwidth}{!}{%
\begin{tabular}{@{}c|cc|cc|c|cc@{}}
\toprule
\multirow{30}{*}{\rotatebox[origin=c]{90}{\textbf{Unibo-powertools}}} & \multicolumn{2}{c|}{\multirow{2}{*}{\textbf{Algorithm}}}                                                                           & \multicolumn{2}{c|}{\textbf{RMSE}} & \multirow{30}{*}{\rotatebox[origin=c]{90}{\textbf{NASA-randomized}}} & \multicolumn{2}{c}{\textbf{RMSE}} \\ \cmidrule(lr){4-5} \cmidrule(l){7-8} 
                                            & \multicolumn{2}{c|}{}                                                                                                              & \textbf{Val}    & \textbf{Test}    &                                            & \textbf{Val}    & \textbf{Test}   \\ \cmidrule(lr){2-5} \cmidrule(l){7-8} 
                                            & \multicolumn{1}{c|}{\multirow{9}{*}{\begin{tabular}[c]{@{}c@{}}Graph Temporal \\ Algorithms(GTAs)\end{tabular}}}     & AGCRN\cite{bai2020adaptive}             & 28.957         & 34.506          &                                            & 44.922         & 38.273          \\
                                            & \multicolumn{1}{c|}{}                                                                                          & DCRNN\cite{li2017diffusion}             & 25.489          & 26.190           &                                            & 26.204          & 26.725          \\
                                            & \multicolumn{1}{c|}{}                                                                                          & DyGrEncoder\cite{taheri2019learning}       & 24.830          & 25.303           &                                            & 22.079          & 22.061          \\
                                            & \multicolumn{1}{c|}{}                                                                                          & GCLSTM\cite{chen2018gc}            & 25.574          & 26.011           &                                            & 23.805          & 24.272          \\
                                            & \multicolumn{1}{c|}{}                                                                                          & GConvGRU\cite{seo2018structured}          & 26.211          & 26.548           &                                            & 23.757          & 23.970          \\
                                            & \multicolumn{1}{c|}{}                                                                                          & GConvLSTM\cite{seo2018structured}         & 26.674          & 27.230           &                                            & 24.691          & 24.784          \\
                                            & \multicolumn{1}{c|}{}                                                                                          & LRGCN\cite{li2019predicting}             & 25.520          & 26.053           &                                            & 24.567          & 24.909          \\
                                            & \multicolumn{1}{c|}{}                                                                                          & MPNN LSTM\cite{panagopoulos2021transfer}          & 37.895          & 37.338           &                                            & 53.197          & 52.344          \\
                                            & \multicolumn{1}{c|}{}                                                                                          & T-GCN\cite{zhao2019t}           & 23.836          & 24.873           &                                            & 24.583          & 24.600          \\ \cmidrule(lr){2-5} \cmidrule(l){7-8} 
                                            & \multicolumn{1}{c|}{\multirow{15}{*}{\begin{tabular}[c]{@{}c@{}}Graph Neural Network \\ Algorithms(GNNs)\end{tabular}}} & AGNN\cite{thekumparampil2018attention}              & 20.743          & 21.058           &                                            & 17.517          & 16.198          \\
                                            & \multicolumn{1}{c|}{}                                                                                          & APPNP\cite{klicpera2018predict}             & 26.909          & 27.081           &                                            & 18.852          & 18.731          \\
                                            & \multicolumn{1}{c|}{}                                                                                          & GatedGraphConv\cite{li2015gated}    & 12.835           & 11.573           &                                            & 12.341          & 10.725          \\
                                            & \multicolumn{1}{c|}{}                                                                                          & GraphConv\cite{morris2019weisfeiler}         & 16.295          & 15.806           &                                            & 15.048          & 14.653          \\
%                                            & \multicolumn{1}{c|}{}                                                                                          & MPNN \cite{gilmer2017neural}                & 0.224           & 0.214            &                                            & 2.317           & 2.384           \\
                                            & \multicolumn{1}{c|}{}                                                                                          & ARMA\cite{bianchi2021graph}              & 9.825           & 9.231            &                                            & 12.405          & 13.961           \\
                                            & \multicolumn{1}{c|}{}                                                                                          & DNA\cite{fey2019just}               & 10.197           & 9.825            &                                            & 12.491           & 13.566           \\
                                            & \multicolumn{1}{c|}{}                                                                                          & GAT\cite{velivckovic2017graph}               & 12.964          & 19.408           &                                            & 13.241          & 11.499          \\
                                            & \multicolumn{1}{c|}{}                                                                                          & GCN\cite{kipf2016semi}               & 13.682          & 14.154           &                                            & 14.086          & 13.759          \\
                                            & \multicolumn{1}{c|}{}                                                                                          & GCN2\cite{chen}              & 17.245          & 16.609           &                                            & 22.561          & 19.326          \\
                                            & \multicolumn{1}{c|}{}                                                                                          & GCChenConv\cite{defferrard2016convolutional}     & 12.110          & 12.310           &                                            & 14.678          & 13.814          \\
                                            & \multicolumn{1}{c|}{}                                                                                          & GraphUNet\cite{gao2019graph}       & 14.126          & 13.865           &                                            & 13.479          & 12.802          \\
                                            & \multicolumn{1}{c|}{}                                                                                          & ResGatedGraphConv\cite{bresson2017residual} & 16.938          & 16.044           &                                            & 32.503          & 21.192          \\
                                            & \multicolumn{1}{c|}{}                                                                                          & SuperGAT\cite{kim2022find}        & 15.835          & 13.543           &                                            & 12.893           & 12.043           \\
                                            & \multicolumn{1}{c|}{}                                                                                          & TAGConv\cite{du2017topology}          & 12.388          & 12.558           &                                            & 14.523          & 14.056          \\ \cmidrule(lr){2-5} \cmidrule(l){7-8} 
                                            & \multicolumn{1}{c|}{\multirow{5}{*}{\begin{tabular}[c]{@{}c@{}}Non-Graph \\ Neural Networks\end{tabular}}}      & GRU\cite{cho2014properties}               & 9.824           & 9.891            &                                            & 12.053           & 11.186           \\
                                            & \multicolumn{1}{c|}{}                                                                                          & LSTM\cite{hochreiter1997long}              & 11.41         & 12.37          &                                            & 13.523         & 11.945          \\
                                            & \multicolumn{1}{c|}{}                                                                                          & Conformer\cite{peng2021conformer}              &    17.83         & 19.57          &                                            &   11.928       &  13.674         \\
                                            & \multicolumn{1}{c|}{}                                                                                          & ETSformer\cite{woo2022etsformer}         & 19.05         & 20.65          &                                            & 23.727          & 21.955          \\ 
                                            & \multicolumn{1}{c|}{}                                                                                          & MoE\cite{baldacchino2016variational}
                                            &     23.61         & 25.37          &                                            &    19.294       &  18.351         
                                            \\
                                            \cmidrule(lr){2-5} \cmidrule(l){7-8} 
                                            & \multicolumn{2}{c|}{\multirow{1}{*}{BGN}}  
                                                    & \textbf{7.800 $\pm$   0.019}      & \textbf{7.570  $\pm$  0.032}       &                                            & \textbf{9.118   $\pm$ 0.013}      & \textbf{8.657    $\pm$  0.045}     
                                            \\ 
                                            \cmidrule(lr){2-5} \cmidrule(l){7-8} 
                                            & \multicolumn{2}{c|}{\multirow{1}{*}{BGN-UE}}  
                                                    & \textbf{7.152  $\pm$  0.035}      & \textbf{7.294   $\pm$  0.041}      &                                            & \textbf{9.827   $\pm$      0.047}  &  \textbf{8.733       $\pm$   0.084}
                                            \\ \bottomrule                                            
\end{tabular}%
}
\end{table}

\vspace{-4mm}
\subsection{Ablation Studies}
\vspace{-2mm}
We perform ablation studies to provide insights into the relative contribution of the Dynamic Graph Inference(DGI) module to the improved overall performance of the BGN model. We model the graph structure learning module with simple algorithmic approaches to design several variants of our proposed framework. We investigate the variant model’s performance compared to the BGN model on all the datasets. We refer to the BGN model for which the DGI module is modeled with the different operators as follows,

\vspace{-2mm}
\begin{itemize}
\item $\textbf{w/} \hspace{1mm} \textbf{fcg}$: \hspace{1mm}$\textbf{BGN}$ model with fully-connected graph(fcg). We disable the DGI module.
\vspace{-1mm}
\item $\textbf{w/o} \hspace{1mm} \mathbf{b}_i$: \hspace{1mm}$\textbf{BGN}$ model with DGI module, for which the sensor embeddings($\mathbf{b}_i$) are not taken into account to infer the dynamic graph. We learn the dynamic graph by leveraging the temporal features($\mathbf{x}^{t}_{i}$) and without the sensor embeddings($\mathbf{b}_i$). We compute the pairwise parameter, i.e., $\theta^{t,k}_{i,j}$ as described below,

\vspace{-2mm}
\resizebox{1\linewidth}{!}{
\begin{minipage}{\linewidth}
\begin{align}
\hspace{20mm}\theta^{t,k}_{i,j} &= \sigma\bigg(g_{fc} \big(\mathbf{W}_{s}\mathbf{x}^{t}_{i} || \mathbf{W}_{s}\mathbf{x}^{t}_{j}\big) \bigg) , \forall j \in\{1, \ldots, n\}, k \in\{0,1\}  \nonumber 
\end{align}
\end{minipage}
}

\vspace{0mm}
\item $\textbf{w/o} \hspace{1mm} \mathbf{x}^{t}_{i}$: \hspace{1mm}$\textbf{BGN}$ model with static graph. We deactivate the DGI module. We learn the static graph by leveraging the sensor embeddings($\mathbf{b}_i$) and without the temporal features($\mathbf{x}^{t}_{i}$). We compute the pairwise parameter, i.e., $\theta^{t,k}_{i,j}$ as described below,

\vspace{-3mm}
\resizebox{1\linewidth}{!}{
\begin{minipage}{\linewidth}
\begin{align}
\hspace{20mm}\theta^{t,k}_{i,j} &= \sigma\bigg(g_{fc} \big(\mathbf{b}_i || \mathbf{b}_j\big) \bigg) , \forall j \in\{1, \ldots, n\}, k \in\{0,1\}  \nonumber 
\end{align}
\end{minipage}
}

\end{itemize}

\vspace{-2mm}
Table \ref{tab:ab} reports the performance of the variant models compared to the BGN model on the test dataset. 

\vspace{-2mm}
\begin{table}[ht!]
\centering
\resizebox{0.95\textwidth}{!}{%
\begin{tabular}{|c|c|c|c|c|c|c|c|c|}
\hline
Algorithms & $\textbf{BGN}$ & $\textbf{w/} \hspace{1mm} \textbf{fcg}$ & $\textbf{w/o} \hspace{1mm} \mathbf{b}_i$ & $\textbf{w/o} \hspace{1mm} \mathbf{x}^{t}_{i}$     \\ \hline
 NASA Data Set\cite{saha2007battery}& \textbf{8.657    $\pm$  0.045} & 17.153    $\pm$  0.195  &  11.735    $\pm$  0.091 &   15.803    $\pm$  0.117   \\ \hline
UNIBO Powertools Dataset\cite{wong2021li} &  \textbf{7.570  $\pm$  0.032} & 16.914  $\pm$  0.102 & 10.973  $\pm$  0.077 &   14.928  $\pm$  0.174  \\ \hline
\end{tabular}%
}
\vspace{2mm}
\caption{\label{tab:ab}The table reports the results of ablation studies.}
\end{table}

\vspace{-5mm}
The RMSE(i.e., prediction error) of the substituted models, $\textbf{w/} \hspace{1mm} \textbf{fcg}$, $\textbf{w/o} \hspace{1mm} \mathbf{b}_i$, $\textbf{w/o} \hspace{1mm} \mathbf{x}^{t}_{i}$, has increased by $\textbf{98.14\%}$, $\textbf{35.55\%}$, $\textbf{82.54\%}$ on NASA Randomized Battery Usage Dataset\cite{saha2007battery}; $\textbf{123.43\%}$, $\textbf{31.01\%}$, $\textbf{97.19\%}$ on the UNIBO Powertools Dataset\cite{wong2021li} compared to the BGN model with relatively low RMSE.
The results show the advantages of utilizing the DGI module, a simple yet effective technique  for capturing the complex structural-dynamic dependencies in the low-dimensional relation graph structure underpinning the battery data. While simultaneously, the GNN backbone provides a useful relational inductive bias for modeling the continuous-time nonlinear dynamics of the complex system to disentangle the various latent aspects underneath the data for better RUL estimates. We conduct additional experiments to investigate the grapher module. We untangle the relative gains of each block in the grapher module by gradually excluding the blocks to design variant models. We then compare the variant model’s performance with the BGN model.

\vspace{-2mm}
\begin{itemize}
\item $\textbf{w/o} \hspace{1mm} \textbf{GNN}$: \hspace{1mm}$\textbf{BGN}$ model without GNN block in grapher module.
\vspace{-1mm}
\item $\textbf{w/o} \hspace{1mm} \mathbf{RNN}$: \hspace{1mm}$\textbf{BGN}$ model without RNN block in grapher module.
\end{itemize}

\vspace{-4mm}
\begin{table}[htbp]
\centering
\resizebox{0.775\textwidth}{!}{%
\begin{tabular}{|c|c|c|c|c|c|c|c|c|}
\hline
Algorithms & $\textbf{BGN}$ & $\textbf{w/o} \hspace{1mm} \textbf{GNN}$ & $\textbf{w/o} \hspace{1mm} \mathbf{RNN}$      \\ \hline
 NASA Data Set\cite{saha2007battery}& \textbf{8.657    $\pm$  0.045} & 13.107    $\pm$  0.067  &  11.913    $\pm$  0.062    \\ \hline
UNIBO Powertools Dataset\cite{wong2021li} &  \textbf{7.570  $\pm$  0.032} & 12.617  $\pm$  0.046 & 10.642  $\pm$  0.074   \\ \hline
\end{tabular}%
}
\vspace{2mm}
\caption{\label{tab:ab1}The table reports the significance of the blocks.}
\end{table}

\vspace{-6mm}
Table \ref{tab:ab1} reports the performance of the variant models without the blocks under investigation compared to the BGN model. The results support the rationale of joint optimization of blocks and demonstrate the competitive performance of our framework on remaining useful life estimation benchmarks. In short, our method leverages the dynamic relation graph learning module and a local graph convolution module coupled with the gating-mechanism neural network to effectively capture the LIBs dynamics for accurate representation modeling yielding improved RUL estimates.

\vspace{-3mm}
\subsection{Hyperparameters Study}

\vspace{-1mm}
\begin{table}[!htbp]
\centering
\caption{Study on model hyperparameter tuning}
\label{tab:hyperparameter}
\resizebox{0.885\textwidth}{!}{%
\begin{tabular}{@{}ccc|c|c|cccccc@{}}
\toprule
\multicolumn{3}{c|}{\multirow{2}{*}{\textbf{Algorithm}}}                                                                                                                                 & \multicolumn{1}{l|}{\multirow{2}{*}{\textbf{MAE}}} & \multicolumn{1}{l|}{\multirow{2}{*}{\textbf{RMSE}}} & \multicolumn{6}{c}{\textbf{Approximation error}}                                                                                                                                           \\ \cmidrule(l){6-11} 
\multicolumn{3}{c|}{}                                                                                                                                                                    & \multicolumn{1}{l|}{}                              & \multicolumn{1}{l|}{}                               & \multicolumn{1}{c|}{\textbf{1}} & \multicolumn{1}{c|}{\textbf{2}} & \multicolumn{1}{c|}{\textbf{3}} & \multicolumn{1}{c|}{\textbf{10}} & \multicolumn{1}{c|}{\textbf{20}} & \textbf{40} \\ \midrule
\multicolumn{1}{c|}{\multirow{16}{*}{\rotatebox[origin=c]{90}{\textbf{Unibo-powertools}}}} & \multicolumn{1}{c|}{\multirow{6}{*}{\textbf{\begin{tabular}[c]{@{}c@{}}Embedded \\ Dimention\end{tabular}}}} & 16     & 10.235                                             & 8.296                                              & \multicolumn{1}{c|}{7.424}     & \multicolumn{1}{c|}{17.061}     & \multicolumn{1}{c|}{23.223}     & \multicolumn{1}{c|}{57.662}      & \multicolumn{1}{c|}{73.776}      & 88.152      \\
\multicolumn{1}{c|}{}                                            & \multicolumn{1}{c|}{}                                                                                        & 32     & 8.206                                              & 7.570                                              & \multicolumn{1}{c|}{9.569}      & \multicolumn{1}{c|}{17.384}     & \multicolumn{1}{c|}{24.083}     & \multicolumn{1}{c|}{56.299}      & \multicolumn{1}{c|}{70.015}      &  87.400      \\
\multicolumn{1}{c|}{}                                            & \multicolumn{1}{c|}{}                                                                                        & 48     & 8.667                                              & 7.488                                              & \multicolumn{1}{c|}{9.713}     & \multicolumn{1}{c|}{19.904}     & \multicolumn{1}{c|}{27.707}     & \multicolumn{1}{c|}{53.981}      & \multicolumn{1}{c|}{69.904}      & 86.465      \\
\multicolumn{1}{c|}{}                                            & \multicolumn{1}{c|}{}                                                                                        & 64     & 9.841                                              & 7.412                                              & \multicolumn{1}{c|}{11.725}      & \multicolumn{1}{c|}{19.401}     & \multicolumn{1}{c|}{26.972}     & \multicolumn{1}{c|}{60.568}      & \multicolumn{1}{c|}{76.656}      & 88.643      \\
\multicolumn{1}{c|}{}                                            & \multicolumn{1}{c|}{}                                                                                        & 96     & 9.891                                              & 7.799                                              & \multicolumn{1}{c|}{10.158}     & \multicolumn{1}{c|}{18.413}     & \multicolumn{1}{c|}{26.349}     & \multicolumn{1}{c|}{58.888}      & \multicolumn{1}{c|}{73.333}      & 85.555      \\
\multicolumn{1}{c|}{}                                            & \multicolumn{1}{c|}{}                                                                                        & 128    & 9.544                                              & 7.618                                              & \multicolumn{1}{c|}{8.399}      & \multicolumn{1}{c|}{15.847}     & \multicolumn{1}{c|}{23.930}     & \multicolumn{1}{c|}{53.566}      & \multicolumn{1}{c|}{71.791}      & 87.322      \\ \cmidrule(l){2-11} 
\multicolumn{1}{c|}{}                                            & \multicolumn{1}{c|}{\multirow{6}{*}{\textbf{\begin{tabular}[c]{@{}c@{}}Hidden \\ Dimention\end{tabular}}}}   & 16     & 13.487                                             & 8.562                                              & \multicolumn{1}{c|}{5.405}      & \multicolumn{1}{c|}{13.195}      & \multicolumn{1}{c|}{18.442}     & \multicolumn{1}{c|}{45.151}      & \multicolumn{1}{c|}{63.911}      & 81.081      \\
\multicolumn{1}{c|}{}                                            & \multicolumn{1}{c|}{}                                                                                        & 32     & 8.206                                              & 7.570                                              & \multicolumn{1}{c|}{9.569}      & \multicolumn{1}{c|}{17.384}     & \multicolumn{1}{c|}{24.083}     & \multicolumn{1}{c|}{56.299}      & \multicolumn{1}{c|}{70.015}      &  87.400      \\
\multicolumn{1}{c|}{}                                            & \multicolumn{1}{c|}{}                                                                                        & 48     & 12.404                                             & 8.178                                              & \multicolumn{1}{c|}{7.643}     & \multicolumn{1}{c|}{15.286}     & \multicolumn{1}{c|}{22.770}     & \multicolumn{1}{c|}{54.140}      & \multicolumn{1}{c|}{70.382}      & 86.146      \\
\multicolumn{1}{c|}{}                                            & \multicolumn{1}{c|}{}                                                                                        & 64     & 8.132                                              & 8.031                                              & \multicolumn{1}{c|}{8.386}     & \multicolumn{1}{c|}{14.873}     & \multicolumn{1}{c|}{21.677}     & \multicolumn{1}{c|}{52.532}      & \multicolumn{1}{c|}{70.569}      & 85.759      \\
\multicolumn{1}{c|}{}                                            & \multicolumn{1}{c|}{}                                                                                        & 96     & 8.693                                              & 6.307                                              & \multicolumn{1}{c|}{12.082}     & \multicolumn{1}{c|}{24.165}     & \multicolumn{1}{c|}{32.750}     & \multicolumn{1}{c|}{63.275}      & \multicolumn{1}{c|}{77.106}      & 89.984      \\
\multicolumn{1}{c|}{}                                            & \multicolumn{1}{c|}{}                                                                                        & 128    & 8.039                                              & 7.140                                              & \multicolumn{1}{c|}{12.400}     & \multicolumn{1}{c|}{24.006}     & \multicolumn{1}{c|}{33.068}     & \multicolumn{1}{c|}{64.228}      & \multicolumn{1}{c|}{77.583}      & 90.938      \\ \cmidrule(l){2-11} 
\multicolumn{1}{c|}{}                                            & \multicolumn{1}{c|}{\multirow{4}{*}{\textbf{\begin{tabular}[c]{@{}c@{}}Learning \\ Rate\end{tabular}}}}      & 0.1    & 24.607                                             & 17.662                                              & \multicolumn{1}{c|}{0.476}      & \multicolumn{1}{c|}{0.793}      & \multicolumn{1}{c|}{1.904}      & \multicolumn{1}{c|}{8.571}      & \multicolumn{1}{c|}{17.619}      & 51.428      \\
\multicolumn{1}{c|}{}                                            & \multicolumn{1}{c|}{}                                                                                        & 0.01   & 10.402                                             & 15.122                                              & \multicolumn{1}{c|}{0.953}      & \multicolumn{1}{c|}{2.543}     & \multicolumn{1}{c|}{3.338}     & \multicolumn{1}{c|}{13.036}      & \multicolumn{1}{c|}{29.888}      & 55.802      \\
\multicolumn{1}{c|}{}                                            & \multicolumn{1}{c|}{}                                                                                        & 0.001  & 8.033                                           & 7.570                                               & \multicolumn{1}{c|}{9.569}      & \multicolumn{1}{c|}{17.384}     & \multicolumn{1}{c|}{24.083}     & \multicolumn{1}{c|}{56.299}      & \multicolumn{1}{c|}{70.015}      &  87.400      \\
\multicolumn{1}{c|}{}                                            & \multicolumn{1}{c|}{}                                                                                        & 0.0001 & 17.986                                             & 10.713                                              & \multicolumn{1}{c|}{3.025}      & \multicolumn{1}{c|}{6.369}     & \multicolumn{1}{c|}{8.757}     & \multicolumn{1}{c|}{33.598}      & \multicolumn{1}{c|}{ 54.777}      & 71.496      \\ \midrule
\multicolumn{1}{c|}{\multirow{16}{*}{\rotatebox[origin=c]{90}{\textbf{NASA-randomized}}}}  & \multicolumn{1}{c|}{\multirow{6}{*}{\textbf{\begin{tabular}[c]{@{}c@{}}Embedded\\ Dimention\end{tabular}}}}  & 16     & 14.329                                             & 8.971                                              & \multicolumn{1}{c|}{29.100}     & \multicolumn{1}{c|}{45.782}     & \multicolumn{1}{c|}{54.692}     & \multicolumn{1}{c|}{65.782}      & \multicolumn{1}{c|}{71.280}      & 82.559      \\
\multicolumn{1}{c|}{}                                            & \multicolumn{1}{c|}{}                                                                                        & 32     & 9.725                                             & 8.657                                              & \multicolumn{1}{c|}{32.796}     & \multicolumn{1}{c|}{45.592}     & \multicolumn{1}{c|}{53.934}     & \multicolumn{1}{c|}{67.014}      & \multicolumn{1}{c|}{74.313}      & 84.645      \\
\multicolumn{1}{c|}{}                                            & \multicolumn{1}{c|}{}                                                                                        & 48     & 14.214                                             & 8.556                                              & \multicolumn{1}{c|}{31.848}     & \multicolumn{1}{c|}{46.919}     & \multicolumn{1}{c|}{54.123}     & \multicolumn{1}{c|}{65.118}      & \multicolumn{1}{c|}{72.701}      & 82.085      \\
\multicolumn{1}{c|}{}                                            & \multicolumn{1}{c|}{}                                                                                        & 64     & 14.703                                             & 9.477                                              & \multicolumn{1}{c|}{28.815}     & \multicolumn{1}{c|}{45.782}     & \multicolumn{1}{c|}{53.839}     & \multicolumn{1}{c|}{65.687}      & \multicolumn{1}{c|}{72.607}      & 82.654      \\
\multicolumn{1}{c|}{}                                            & \multicolumn{1}{c|}{}                                                                                        & 96     & 12.421                                             & 7.923                                              & \multicolumn{1}{c|}{30.047}     & \multicolumn{1}{c|}{45.687}     & \multicolumn{1}{c|}{53.555}     & \multicolumn{1}{c|}{66.825}      & \multicolumn{1}{c|}{74.028}      & 85.782      \\
\multicolumn{1}{c|}{}                                            & \multicolumn{1}{c|}{}                                                                                        & 128    & 14.384                                             & 9.117                                              & \multicolumn{1}{c|}{30.237}     & \multicolumn{1}{c|}{46.256}     & \multicolumn{1}{c|}{54.408}     & \multicolumn{1}{c|}{66.540}      & \multicolumn{1}{c|}{72.891}      & 82.180      \\ \cmidrule(l){2-11} 
\multicolumn{1}{c|}{}                                            & \multicolumn{1}{c|}{\multirow{6}{*}{\textbf{\begin{tabular}[c]{@{}c@{}}Hidden \\ Dimention\end{tabular}}}}   & 16     & 14.991                                             & 9.125                                              & \multicolumn{1}{c|}{27.299}     & \multicolumn{1}{c|}{42.275}     & \multicolumn{1}{c|}{50.427}     & \multicolumn{1}{c|}{66.066}      & \multicolumn{1}{c|}{73.270}      & 82.085      \\
\multicolumn{1}{c|}{}                                            & \multicolumn{1}{c|}{}                                                                                        & 32     & 9.476                                             & 8.657                                              & \multicolumn{1}{c|}{32.796}     & \multicolumn{1}{c|}{45.592}     & \multicolumn{1}{c|}{53.934}     & \multicolumn{1}{c|}{67.014}      & \multicolumn{1}{c|}{74.313}      & 84.645      \\
\multicolumn{1}{c|}{}                                            & \multicolumn{1}{c|}{}                                                                                        & 48     & 14.363                                             & 9.439                                              & \multicolumn{1}{c|}{29.479}     & \multicolumn{1}{c|}{45.877}     & \multicolumn{1}{c|}{55.071}     & \multicolumn{1}{c|}{66.919}      & \multicolumn{1}{c|}{73.555}      & 82.938      \\
\multicolumn{1}{c|}{}                                            & \multicolumn{1}{c|}{}                                                                                        & 64     & 14.922                                             &  8.863                                              & \multicolumn{1}{c|}{34.123}     & \multicolumn{1}{c|}{49.005}     & \multicolumn{1}{c|}{55.545}     & \multicolumn{1}{c|}{67.867}      & \multicolumn{1}{c|}{72.796}      & 81.611      \\
\multicolumn{1}{c|}{}                                            & \multicolumn{1}{c|}{}                                                                                        & 96     & 12.282                                             & 8.977                                              & \multicolumn{1}{c|}{35.450}     & \multicolumn{1}{c|}{50.332}     & \multicolumn{1}{c|}{57.251}     & \multicolumn{1}{c|}{69.194}      & \multicolumn{1}{c|}{75.640}      & 85.782      \\
\multicolumn{1}{c|}{}                                            & \multicolumn{1}{c|}{}                                                                                        & 128    & 11.736                                             & 8.233                                              & \multicolumn{1}{c|}{38.199}     & \multicolumn{1}{c|}{52.607}     & \multicolumn{1}{c|}{58.294}     & \multicolumn{1}{c|}{68.057}      & \multicolumn{1}{c|}{75.829}      & 86.351      \\ \cmidrule(l){2-11} 
\multicolumn{1}{c|}{}                                            & \multicolumn{1}{c|}{\multirow{4}{*}{\textbf{\begin{tabular}[c]{@{}c@{}}Learning \\ Rate\end{tabular}}}}      & 0.1    & 31.478                                             & 43.634                                              & \multicolumn{1}{c|}{17.156}      & \multicolumn{1}{c|}{28.436}     & \multicolumn{1}{c|}{38.199}     & \multicolumn{1}{c|}{64.455}      & \multicolumn{1}{c|}{71.469}      & 80.947      \\
\multicolumn{1}{c|}{}                                            & \multicolumn{1}{c|}{}                                                                                        & 0.01   & 12.831                                             & 15.252                                              & \multicolumn{1}{c|}{13.364}     & \multicolumn{1}{c|}{26.351}     & \multicolumn{1}{c|}{38.767}     & \multicolumn{1}{c|}{64.834}      & \multicolumn{1}{c|}{74.502}      & 86.066      \\
\multicolumn{1}{c|}{}                                            & \multicolumn{1}{c|}{}                                                                                        & 0.001  & 9.935                                             & 9.193                                              & \multicolumn{1}{c|}{32.796}     & \multicolumn{1}{c|}{45.592}     & \multicolumn{1}{c|}{53.934}     & \multicolumn{1}{c|}{67.014}      & \multicolumn{1}{c|}{74.313}      & 84.645      \\
\multicolumn{1}{c|}{}                                            & \multicolumn{1}{c|}{}                                                                                        & 0.0001 & 17.754                                             & 12.612                                              & \multicolumn{1}{c|}{16.208}     & \multicolumn{1}{c|}{32.227}     & \multicolumn{1}{c|}{39.336}     & \multicolumn{1}{c|}{60.853}      & \multicolumn{1}{c|}{69.668}      & 79.431      \\ \bottomrule
\end{tabular}%
}
\end{table} 
\vspace{-1mm}
We conduct hyperparameter tuning using the grid search technique to determine the optimal hyperparameter values. We perform hyperparameters optimization to minimize the RMSE on the validation dataset.  The training hyperparameter configuration was explored from the following ranges: node embedding dimension(for $\mathbf{z}_i$) $\in [16, 128]$, the hidden dimension(for $\mathbf{h}_i$)$\in [16, 128]$, the learning rate(lr) $\in [0.1, 0.0001]$. The selected hyperparameter ranges are the same across NASA Randomized Battery Usage Dataset\cite{saha2007battery} and UNIBO Powertools Dataset\cite{wong2021li}, BGN models, and experimental settings. Table \ref{tab:hyperparameter} reports the results of the hyperparameter study in terms of performance evaluation metrics, RMSE and MAE. We also evaluate the model performance in terms of approximation error. It is the difference in the actual value and estimated value compared to the actual value in terms of percentage. Suppose that for a given predefined approximation error,  the result value corresponding to it is the percentage of time steps in the dataset whose model prediction error is less than the predefined approximation error. The results suggest that our model is sensitive to the choice of hyperparameters values. The optimal values determined from the hyperparameter sensitivity analysis of the best performing model are as follows, node embedding dimension and the hidden dimension equal to 32, and the optimal value of lr is 0.001. 

\vspace{-4mm}
\section{Conclusion}
\vspace{-3mm}
Precise battery remaining useful life estimation is crucial for battery-powered systems. We present the Battery GraphNets(BGN) framework that jointly learns the discrete dependency structures of the battery parameters and exploits the relational dependencies through neural message-passing schemes to provide accurate RUL estimates. We achieve SOTA performance on the open-source benchmark battery datasets on the RUL prediction task. The experimental results support the effectiveness of our approach for achieving better performance compared to the state-of-art methods.

\vspace{-3mm} %  to distil the knowledge gained by our proposed method into a succinct form
\section*{Broader Impact}  % Causal discovery in time-varying dynamical settings; Identification and estimation of causal effects in dynamical systems
\vspace{-3mm}
The current techniques in literature for prognostics or the remaining service life estimation of batteries are classified as follows, (a) physics-based, and (b) data-driven modeling techniques. The traditional physics-based modeling methods rely on the first-principles approach-based techniques to develop LiBs equivalent models. The physics-aware mechanistic models have inherent drawbacks of the high computational complexity of numerical simulations. The high-fidelity Reduced-Order Electrochemical Models (ROECMs) and simulations for LiBs are computationally tractable. However, they suffer from poor uncertainty estimates and causal inference mechanisms for precise RUL estimates. The existing data-driven techniques neglect the inter-relations among multiple battery parameters in modeling the complex battery degradation trajectories. We utilize the emerging first-class citizens of deep learning, the GNNs, for encoding the relational dependencies between battery parameters to provide high-quality RUL estimates. There is immense scope for hybrid models in which the graph deep learning-based battery life estimation models are complemented by the mechanistic models to further improve the Battery Management Systems(BMS) technology for improved degradation prognostics of LiBs.

\appendix
\vspace{-5mm}
\section{Appendix}
\vspace{-3mm}
\subsection{Graph Contrastive Algorithms}  
\vspace{-1mm}
Furthermore, we compare the performance of the graph self-supervised techniques, i.e., graph-contrastive learning algorithms(GCL, \cite{Zhu:2021tu}) on the RUL estimation task w.r.t, to the BGN model trained in supervised settings. The GCL algorithms are classified based on three main components, (a) graph augmentation techniques, (b) contrasting architectures and modes, and (c) contrastive objectives. The graph-input-based GCL algorithms utilize the DGI module to learn the dynamic graph structure. Here, we briefly discuss the baseline GCL techniques,

\vspace{-2mm}
\begin{itemize}
\item $\text{BGRL}$\cite{thakoor2021bootstrapped}: a) utilizes edge removing and feature masking augmentation techniques, (b) the contrastive objective is the Bootstrapping Latent(BL) loss. (c) The contrastive modes are cross-scale(G2L, i.e., Global-Local mode) and same-scale contrasting(L2L, i.e., Local-Local mode), and (c) utilizes bootstrapped contrasting technique.

Note: L2L and G2G contrasts the embedding pairs of the same node/graph with different views as positive pairs. G2L contrast the cross-scale, i.e., the node-graph embedding pairs at differing granularity as positive pairs.

\item GBT\cite{bielak2021graph}: (a) utilizes edge removing and feature masking augmentation techniques, (b) the contrastive objective is the Barlow Twins (BT) loss.
(c) the contrastive modes are Local-Local (L2L), Global-Global (G2G), and leverage the within-embedding contrasting technique.
\item GRACE\cite{zhu2020deep}: a) utilizes edge removing and feature masking augmentation techniques, (b) the contrastive objective is the InfoNCE loss. (c) the contrastive mode is Local-Local(L2L), and (c) utilizes a dual-branch contrasting technique.
\item GraphCL\cite{you2020graph}: a) utilizes subgraphs induced by random walks (RWS), node dropping, edge removing, and feature masking augmentation techniques, (b) the contrastive objective is the InfoNCE loss, the contrastive mode is Global-Global (G2G) and (c) utilizes a dual-branch contrasting technique.
\item InfoGraph\cite{sun2019infograph}: (a) the contrastive objective is the Jensen-Shannon Divergence (JSD) loss. (b) The contrastive mode is Global-Local(G2L), and (c) utilizes a single-branch contrasting technique.
\item MVGRL\cite{hassani2020contrastive}: a) utilizes the Personalized PageRank(PPR), Markov Diffusion Kernel(MDK) augmentation techniques, (b) the contrastive objective is the Jensen-Shannon Divergence(JSD) loss,  the contrastive mode is Global-Local(G2L), and (c) utilizes dual-branch contrasting technique. 
\end{itemize}

\vspace{-5mm}
\begin{table}[!htbp]
\centering
\caption{Comparative study of the GCL algorithms on RUL estimation task.}
\vspace{1mm}
\label{tab:GCL}
\setlength{\tabcolsep}{4pt}
\resizebox{1\textwidth}{!}{%
\begin{tabular}{@{}c|c|cc|cc|cc|cc|cc@{}}
\toprule
\multirow{2}{*}{\textbf{Dataset}}                                                     & \multirow{2}{*}{\textbf{\begin{tabular}[c]{@{}c@{}}Graph-Contrastive\\ Learning\end{tabular}}} & \multicolumn{2}{c|}{\textbf{SVM}} & \multicolumn{2}{c|}{\textbf{RF}} & \multicolumn{2}{c|}{\textbf{GradientBoosting}} & \multicolumn{2}{c|}{\textbf{AdaBoost}} & \multicolumn{2}{c}{\textbf{MLPRegressor}} \\ \cmidrule(l){3-12} 
                                                                                      &                                                                                                & \textbf{MAE}    & \textbf{MSQR}   & \textbf{MAE}   & \textbf{MSQR}   & \textbf{MAE}          & \textbf{MSQR}          & \textbf{MAE}      & \textbf{MSQR}      & \textbf{MAE}        & \textbf{MSQR}       \\ \midrule
\multirow{7}{*}{\rotatebox[origin=c]{90}{\textbf{\begin{tabular}[c]{@{}c@{}}Unibo-\\ powertools\end{tabular}}}} & BGRL\_G2L\cite{thakoor2021bootstrapped}                                                                                       & 70.08           & 7344.82         & 52.80          & 6314.80         & 54.35                 & 6682.64                & 58.70             & 6904.74            & 56.84               & 6897.95             \\
                                                                                      & BGRL\_L2L\cite{thakoor2021bootstrapped}                                                                                       & 58.22           & 6653.74         & 51.11          & 6086.23         & 51.80                 & 6213.19                & 56.52             & 6587.25            & 55.26               & 6329.98             \\
                                                                                      & GBT\cite{bielak2021graph}                                                                                            & 31.03           & 1481.81         & 22.50          & 1932.37         & 30.25                 & 2511.33                & 76.61             & 7493.91            & 12.15               & 945.23              \\
                                                                                      & GRACE\cite{zhu2020deep}                                                                                          & 32.62           & 1692.85         & 9.42           & 549.30          & 18.14                 & 1037.32                & 63.45             & 5315.61            & 21.65               & 1438.78             \\
                                                                                      & GraphCL\cite{you2020graph}                                                                                        & 41.93           & 3452.56         & 20.66          & 2595.74         & 31.88                 & 3308.71                & 48.06             & 4798.50            & 28.09               & 2793.75             \\
                                                                                      & InfoGraph\cite{sun2019infograph}                                                                                      & 42.50           & 3617.82         & 23.79          & 2771.81         & 33.88                 & 3378.53                & 51.73             & 4525.64            & 28.62               & 2728.35             \\
                                                                                      & MVGRL\cite{hassani2020contrastive}                                                                                   & 35.05           & 2349.90         & 19.04          & 1910.63         & 26.58                 & 2419.41                & 52.66             & 4647.54            & 29.46               & 2537.30             \\ \midrule
\multirow{7}{*}{\rotatebox[origin=c]{90}{\textbf{\begin{tabular}[c]{@{}c@{}}NASA-\\ randomized\end{tabular}}}}  & BGRL\_G2L                                                                                     \cite{thakoor2021bootstrapped}                                                                                       & 65.16           & 6093.01         & 64.71          & 6158.46         & 64.56                 & 5847.53                & 74.06             & 6800.71            & 67.22               & 6103.45             \\
                                                                                      & BGRL\_L2L                                                                                     \cite{thakoor2021bootstrapped}                                                                                       & 65.28           & 6185.40         & 63.47          & 5737.31         & 64.07                 & 5944.42                & 76.22             & 7035.51            & 66.22               & 6200.28             \\
                                                                                      & GBT\cite{bielak2021graph}                                                                                                                                                                                       & 34.89           & 1557.71         & 18.99          & 953.75          & 22.40                 & 1028.03                & 56.33             & 4197.77            & 15.61               & 614.94              \\
                                                                                      & GRACE\cite{zhu2020deep}                                                                                        & 26.97           & 1075.23         & 5.09           & 180.94          & 9.38                  & 240.00                 & 53.82             & 3676.08            & 23.38               & 1017.55             \\
                                                                                      & GraphCL\cite{you2020graph}                                                                                        & 30.98           & 1537.57         & 15.26          & 591.94          & 22.21                 & 923.43                 & 58.34             & 4388.31            & 19.76               & 796.15              \\
                                                                                      & InfoGraph\cite{sun2019infograph}                                                                                      & 37.12           & 2160.39         & 27.58          & 1700.42         & 35.22                 & 2162.31                & 58.54             & 4665.66            & 30.80               & 1821.47             \\
                                                                                      & MVGRL\cite{hassani2020contrastive}                                                                                                                                                                     & 33.69           & 1539.46         & 23.03          & 1152.42         & 31.88                 & 1831.67                & 72.49             & 6446.51            & 30.79               & 1711.52             \\ \bottomrule
\end{tabular}%
}
\end{table}

\vspace{1mm}
In Table \ref{tab:GCL}, we report the results obtained from Graph-Contrastive Learning Algorithms(GCL, \cite{Zhu:2021tu}) on the RUL prediction task. The GCL algorithms learn the unsupervised entire graph representations. The node-level graph encoder of the GCL algorithms is modeled with GCN\cite{kipf2016semi} to obtain unsupervised node representations. We performed global average pooling of node representations to compute the unsupervised graph-level representation. We utilize several ML techniques, such as Support Vector Machine($\text{SVM}$, \cite{cortes1995support}), Random Forest($\text{RF}$, \cite{liaw2002classification}), $\text{GradientBoosting}$\cite{friedman2001greedy}, $\text{AdaBoost}$\cite{schapire2013explaining} \& $\text{MLP}$\cite{haykin1994neural} trained in supervised settings to model the RUL estimates as a function of the unsupervised graph-level representations. We evaluate the ML model's performance on the test dataset to report the quality of unsupervised graph-level representations in terms of RUL prediction error reported in $\text{MAE}$ \& $\text{MSQR}$. Our proposed method, i.e., the BGN model trained through a supervised learning approach,  reports a lower MAE error of $9.768 \pm 0.083$ on the NASA Dataset\cite{saha2007battery}; $8.291 \pm 0.054$ on the UNIBO Dataset\cite{wong2021li}. We observe a $\textbf{63.78\%}$ on the NASA Dataset\cite{saha2007battery} and; $\textbf{73.27\%}$ on the UNIBO Dataset\cite{wong2021li} lower MAE error compared to the next-best GCL algorithm trained through self-supervised learning.

\vspace{-2mm}
\subsection{Baseline Algorithms}
\vspace{-2mm}
The graph-input-based baseline algorithms utilize the DGI module to learn the dynamic graph structure. Here, we discuss the short description of the baseline GCN algorithms. 

\vspace{-3mm}
\begin{itemize}
\item AGNN\cite{thekumparampil2018attention} : The algorithm presents a neural-message passing schema to obtain the whole graph representation.  It computes the cosine similarity of the node embeddings to obtain the propagation matrix that accounts for the dynamic and adaptive summary of the local-graph neighborhood.  We utilize the propagation matrix to refine the node embeddings. 

\item GCN\cite{kipf2016semi} : The algorithm is a vanilla variant of the convolutional neural network operator to encode the discrete graphs to low-dimensional embeddings. It aggregates the normalized neural messages from the local neighborhood to update each node's fixed-length embeddings.                                                                                                                   
\item GCN2\cite{chen} : The algorithm extends the vanilla GCN operator. It integrates the initial residual connection and identity mapping strategy to overcome the over-smoothing issue.

\item APPNP\cite{klicpera2018predict} :  The algorithm presents an improved neural-message passing schema based on the personalized PageRank technique. The model trained end-to-end by the gradient-based optimization technique utilizes the initial-connection strategy to refine the node embeddings for learning on graphs.

\item  GatedGraphConv\cite{li2015gated}: The algorithm presents node-update equations. It utilizes the GRU unit to refine the node embeddings by regulating the information flow aggregated from the local-graph neighbors and outputs the transformed node embedding sequences. 

\item GraphConv\cite{morris2019weisfeiler} : The algorithm presents the k-dimensional GNNs that account for the fine- and coarse-grained structures to compute the graph representations at multiple scales for better inference of the graph-structured data.
 
\item   ARMA\cite{bianchi2021graph} : The algorithm substitutes the conventional  polynomial spectral filters with the auto-regressive moving average (ARMA) filters to perform convolution on the graphs to better capture the global graph structure.       
  
\item  DNA\cite{fey2019just} :  The algorithm presents the dynamic neighborhood aggregation (DNA) scheme to perform a node-adaptive aggregation of neighboring embeddings. This algorithmic approach offers a highly-dynamic receptive field to preserve the higher-order features of the dominant subgraph structures.

\item  GAT\cite{velivckovic2017graph} :  The algorithm presents the local-graph attention strategy to refine the node-level representations by weighing the local-graph neighbors of importance. Stacking multiple-layers aggregates high-order information to compute the expressive node embeddings.
                            
\item  GCChenConv\cite{defferrard2016convolutional} : The algorithm utilizes the fast localized spectral filtering from spectral graph theory to design low-pass convolutional filters on graphs and coupled with a graph-coarsening and pooling operation for higher filter resolution.
                                                                                                                            
\item GraphUNet\cite{gao2019graph}  : The algorithm is an encoder-decoder architecture to operate on arbitrary-sized graphs. It presents the differentiable local-graph pooling and unpooling operations for better capturing the hierarchical information to encode the dominant structural characteristics of the discrete graphs.

\item ResGatedGraphConv\cite{bresson2017residual} : The algorithm presents the generic formulation of residual gated graph convolutional operator that integrates the graph-based RNN and ConvNet architectures to operate on discrete graph-structured data with arbitrary sizes.
                                                                                                                                          
\item SuperGAT\cite{kim2022find} : The algorithm presents the self-supervised graph attention network. It performs the self-supervised attention through the novel graph neural network architecture design and generalizes well on the graphs with arbitrary sizes.
    
\item  TAGConv\cite{du2017topology} : The algorithm presents the topology-adaptive, graph convolutional network operator that learns the fixed-size learnable filters adaptive to arbitrary-sized discrete graph-structured data. 
\end{itemize} 
 
\vspace{-2mm} 
Now, we discuss the short description of the baseline Graph Temporal algorithms. 
 
\vspace{-2mm}
\begin{itemize}

\item AGCRN\cite{bai2020adaptive} : The graph convolutional recurrent network algorithm learns the fine-grained spatial and temporal dependencies in the correlated time series data by simultaneously encoding the inter- and intra- dependencies among multiple independent time series automatically in the graph representations.
   
\item DCRNN\cite{li2017diffusion} :  The algorithm presents the encoder-decoder architecture with the scheduled sampling technique. It integrates the local-graph diffusion convolution operator and recurrent architectures to capture the complex spatial dependencies and the non-linear temporal dynamics.

\item DyGrEncoder\cite{taheri2019learning} : The algorithm presents the unsupervised graph-representation learning method. It is based on encoder-decoder architecture to learn the topological and temporal features of the time-evolving dynamic graphs. 

\item GCLSTM\cite{chen2018gc} : The algorithm couples the graph convolution network(GCN) to identify spatial structures and the long short-term memory network (LTSM)
to learn the dynamic patterns of the discrete time-varying graphs. 

\item GConvGRU\cite{seo2018structured} : The algorithm presents the graph convolutional recurrent networks that couples the ConvNets and gated recurrent network(GRU) to perform inference on graphs. It jointly learns  the  local-node structure and temporal features. The novel neural network architecture exploits graph-spatial and the dynamic feature information for learning on time-varying graphs.

\item LRGCN\cite{li2019predicting}  : The algorithm captures both the temporal dependency and graph structure dynamics. The graph convolution operator jointly learns  intra- and inter-time dependencies of time-dependent discrete graphs.

\item MPNN LSTM\cite{panagopoulos2021transfer} : The algorithm presents a representation learning technique on graphs to learn the complex topological structures and the temporal dependencies for better learning of the spatio-temporal dynamics between the time-adjacent discrete graph snapshots.

\item T-GCN\cite{zhao2019t}:  The algorithm captures the spatial dependence with the graph convolutional network (GCN), and the gated recurrent unit learns the dynamic changes of the time-evolving discrete graphs.
 
\end{itemize}

\vspace{-4mm}
\subsection{Additional Experiments}

\vspace{0mm}
\subsubsection{Data Augmentation for Battery Life Estimation in Battery Digital Twins}
\vspace{-1mm} 
The objective of the data augmentation is two-fold. The first objective is to generate a synthetic-battery dataset while conserving probability distributions and spatio-temporal dynamics of the real-battery dataset. In this work context, the real-battery dataset is the battery parameters observations. The synthetic-battery dataset generation potentially serves as virtual simulations to capture the battery performance, an alternative research paradigm to first-principles-based non-linear-mechanistic models and numerical solutions. The second objective is to demonstrate the lower prediction error of the baseline prediction model trained jointly with the real-battery training dataset and the synthetic-battery dataset on the real-holdout set compared to the variant of the baseline prediction model trained with the real-battery training dataset to estimate RUL. We demonstrate the graph deep learning-based Variational Auto-Encoders (VAEs, \cite{kingma2013auto}) efficacy on synthetic dataset generation and its utility in the downstream task in predicting RUL on real-world open-source NASA Randomized Battery Usage Dataset\cite{saha2007battery} and the UNIBO Powertools Dataset\cite{wong2021li}.

\vspace{1mm}
\textbf{Framework:} 
The VAE-based generative model consists of (a) inference neural network $g_\phi(\mathbf{x}^{t}_{i})$ and (b) generative neural network $g_\theta(\mathbf{z}^{t}_{i})$. The encoding distribution of the probabilistic encoder $g_\phi(\mathbf{x}^{t}_{i})$ is given as $p(\mathbf{z}^{t}_{i}\vert\mathbf{x}^{t}_{i})$ and approximated as $q_\phi(\mathbf{z}^{t}_{i}\vert\mathbf{x}^{t}_{i})$. The probabilistic encoder $g_\phi(\mathbf{x}^{t}_{i})$ computes latent variables $\mathbf{z}^{t}_{i}$ given observed data $\mathbf{x}^{t}_{i}$ by approximating the true posterior distribution $p(\mathbf{z}^{t}_{i}\vert\mathbf{x}^{t}_{i})$ as $q_\phi(\mathbf{z}^{t}_{i}\vert\mathbf{x}^{t}_{i})$. The decoding distribution of the probabilistic decoder $g_\theta(\mathbf{z}^{t}_{i})$ is denoted by $p(\mathbf{x}^{t}_{i}\vert\mathbf{z}^{t}_{i})$ and approximated as $p_\theta(\mathbf{x}^{t}_{i}\vert\mathbf{z}^{t}_{i})$. The probabilistic decoder $g_\theta(\mathbf{z}^{t}_{i})$ maps latent variable $\mathbf{z}^{t}_{i}$ to $\mathbf{x}^{t}_{i}$ by approximating true likelihood $p(\mathbf{x}^{t}_{i}\vert\mathbf{z}^{t}_{i})$ as $p_\theta(\mathbf{x}^{t}_{i}\vert\mathbf{z}^{t}_{i})$. We sample the latent variables $\mathbf{z}^{t}_{i}$ from a prior distribution, i.e., a standard normal distribution. We optimize the parameters $\phi, \theta$ of the encoding network $q_\phi(\mathbf{z}^{t}_{i}\vert\mathbf{x}^{t}_{i})$ and the decoding network $p_\theta(\mathbf{x}^{t}_{i}\vert\mathbf{z}^{t}_{i})$ by maximizing the variational lower bound of the probability of sampling real-battery dataset as described below,

\vspace{-3mm}
\resizebox{1.05\linewidth}{!}{
\begin{minipage}{\linewidth}
\begin{align}
L_\text{VAE}(\theta, \phi) 
&= -\log p(\mathbf{x}^{t}_{i}) + D_\text{KL}( q_\phi(\mathbf{z}^{t}_{i}\vert\mathbf{x}^{t}_{i}) \| p(\mathbf{z}^{t}_{i}\vert\mathbf{x}^{t}_{i}) ) \nonumber \\ 
&= - \mathbb{E}_{\mathbf{z}^{t}_{i} \sim q_\phi(\mathbf{z}^{t}_{i}\vert\mathbf{x}^{t}_{i})}\log p_\theta(\mathbf{x}^{t}_{i}\vert\mathbf{z}^{t}_{i}) + D_\text{KL}(q_\phi(\mathbf{z}^{t}_{i}\vert\mathbf{x}^{t}_{i}) \| p(\mathbf{z}^{t}_{i})) \nonumber \\ 
\theta^{*}, \phi^{*} &= \arg\min_{\theta, \phi} L_\text{VAE} \nonumber
\end{align}
\end{minipage}
}

\vspace{-1mm}
The VAE model (a) maximizes the (log-)likelihood of generating a real-battery dataset, i.e., $\log p(\mathbf{x}^{t}_{i})$ when learning the true-probability distribution. (b) minimizes the difference between the estimated posterior, i.e., $q_\phi(\mathbf{z}^{t}_{i}\vert\mathbf{x}^{t}_{i})$ and real $p(\mathbf{z}^{t}_{i}\vert\mathbf{x}^{t}_{i})$. (c) minimizes the reconstruction loss, i.e., $p_\theta(\mathbf{x}^{t}_{i}\vert\mathbf{z}^{t}_{i})$, which incentivizes the decoder network to effectively transform the latent variables $\mathbf{z}^{t}_{i}$ to $\mathbf{x}^{t}_{i}$. (d) minimizes the KL divergence  between real prior $p(\mathbf{z}^{t}_{i})$ and estimated posterior distribution, $q_\phi(\mathbf{z}^{t}_{i}\vert\mathbf{x}^{t}_{i})$.

\vspace{1mm}
\textbf{Encoder:}  
We model the probabilistic encoder with the BGN model. The input to the encoder is the real-battery data $\mathbf{x}^{t}_{i}$. The DGI module learns the adjacency matrix $\mathcal{A}^{t}$ at time-point t. The grapher module computes the node-level representations $\mathbf{h}_i^{t}$. We optimize the node embeddings $\mathbf{b}_i$ along with the other training parameters of the VAE. The pooling operator performs the element-wise product of $\mathbf{h}_i^{t}$ and $\mathbf{b}_i$. It then performs mean average pooling to obtain graph-level representation. We linearly transform the entire graph representation to compute the mean and variance of the distribution, $\mu^{t}_{i}$, and $\sigma^{t}_{i}$, respectively. We obtain the latent variable $z^{t}_{i}$ through the reparameterization trick as described below,

\vspace{-5mm}
\resizebox{1\linewidth}{!}{
\begin{minipage}{\linewidth}
\begin{align}
z^{t}_{i} &\sim q_\phi(z^{t}_{i}\vert x^{t}_{i}, \mathcal{A}^{t}) = \mathcal{N}(z^{t}_{i}; \boldsymbol{\mu}^{t}_{i}, (\boldsymbol{\sigma}^{t}_{i})^{2}\boldsymbol{I}) & \\
z^{t}_{i} &= \boldsymbol{\mu}^{t}_{i} + \boldsymbol{\sigma}^{t}_{i} \odot \boldsymbol{\epsilon} \text{, where } \boldsymbol{\epsilon} \sim \mathcal{N}(0, \boldsymbol{I}) & 
\end{align}
\end{minipage}
}

\vspace{-1mm}
\textbf{Decoder:}  
We model the probabilistic decoder with the BGN neural network and have different trainable parameters compared to the encoder. The input to the decoder is the latent variable $z^{t}_{i}$. The DGI module reconstructs the adjacency matrix, $\tilde{\mathcal{A}}^{t}$ at time-point t. The grapher module learns the reconstructed node-level representations $\tilde{\mathbf{h}}_i^{t}$. The pooling operator computes the element-wise product of $\tilde{\mathbf{h}}_i^{t}$, $\mathbf{b}_i$ and outputs the reconstructed real-battery dataset, $\tilde{\mathbf{x}}^{(t)}_{i}$. We additionally reduce the binary cross entropy loss between the real $\mathcal{A}^{t}$ and reconstructed $\tilde{\mathcal{A}}^{t}$. After training the VAE model on $\mathbf{x}^{t}_{i}$, the synthetic-battery data $\tilde{\mathbf{x}}^{t}_{i}$ is determined independently by sampling sequences using the generative decoder.

\vspace{1mm}
\textbf{RUL Prediction :}
We demonstrate the benefits of the synthetic-battery dataset. We perform experiments on the NASA Randomized Battery Usage Dataset\cite{saha2007battery} and the UNIBO Powertools Dataset\cite{wong2021li}. We train the VAE algorithm on the real-battery training dataset. We utilized the validation dataset for hyperparameter tuning and tuned model selection. We sampled the synthetic data from the VAE framework. We select the BGN framework as a baseline model trained by the real-battery training dataset for RUL prediction. The BGN$^*$ denotes the baseline model variant trained jointly with the real-battery training and the sampled synthetic-battery dataset. We evaluate the performance of both models on the real-battery holdout dataset. We report in \autoref{tab:syntheticdata} a $\textbf{60.65}\%$ \& $\textbf{46.47}\%$ drop in prediction error(RMSE) of BGN$^*$ model performance on NASA Randomized Battery\cite{saha2007battery} and UNIBO \cite{wong2021li} holdout datasets, respectively. The synthetic-battery dataset has learned key-dominant patterns across the real-battery training dataset and is generalized better. It leads to better performance of the BGN$^*$ model on the real-battery holdout set compared to the BGN baseline model.

\vspace{-2mm}
\begin{table}[htbp]
\centering
\resizebox{0.6\textwidth}{!}{%
\begin{tabular}{|c|c|c|c|c|c|c|c|c|}
\hline
Algorithms & $\textbf{BGN}$ & $\textbf{BGN}^*$    \\ \hline
 NASA Data Set\cite{saha2007battery}& \textbf{8.657    $\pm$  0.045} & \textbf{3.406  $\pm$  0.193}    \\ \hline
UNIBO Powertools Dataset\cite{wong2021li} &  \textbf{7.570  $\pm$  0.032}  & \textbf{4.052  $\pm$  0.281} \\ \hline
\end{tabular}%
}
\vspace{1mm}
\caption{\label{tab:syntheticdata}Performance comparison of models on RUL prediction.}
\end{table}

\vspace{-6mm}
\subsection{Missing Battery Data Imputation for Battery Life Estimation in Battery Digital Twins}
\vspace{-2mm} 
The objective of missing data imputation is two-fold. The first objective is to impute missing data of battery parameters. The second objective is to demonstrate the utility of the imputed data. The algorithmic framework is driven by the two-player, non-cooperative, zero-sum adversarial game based on game theory and a minimax optimization approach. The framework with an adversarially trained generator neural network presents an alternative paradigm to numerical modeling and simulations of first principles based on imputation techniques. We leverage artificial intelligence systems to demonstrate the random missing data imputation-utility efficacy tradeoff for the downstream task of RUL prediction on real-world open-source NASA Randomized Battery Usage Dataset\cite{saha2007battery} and the UNIBO Powertools Dataset\cite{wong2021li}. 

\textbf{Framework:}
The Wasserstein Generative Adversarial Network(WGAN, \cite{arjovsky2017wasserstein}) is trained to impute the unobserved data of $\mathbf{x}^{t}_{i}$.  
Let us assume that $\hat{\mathbf{x}}^{t}_{i}$ denotes the imputed dataset. The generator network of the WGAN framework learns a density $\hat{p}(\hat{\mathbf{x}}^{t}_{i})$  of $\hat{\mathbf{x}}^{t}_{i}$ that best approximates the probability density function, $\small p(\mathbf{x}^{t}_{i})$ of $\mathbf{x}^{t}_{i}$. The mathematical description is as follows,

\vspace{-3mm} 
\begin{equation}
\small \min _{\hat{p}} \mathcal{W} \big(p(\mathbf{x}^{t}_{i}) \|  \hspace{1mm} \hat{p}(\hat{\mathbf{x}}^{t}_{i})\big)
\end{equation}

\vspace{-2mm} 
$\mathcal{W}$ denotes the Wasserstein distance of order-1. In general, $\hat{\mathbf{x}}^{t}_{i}$ had obtained by determining the Nash equilibrium of the competing game-setting of two distinct machine players, $\mathcal{G}_{n}$ and $\mathcal{D}_{n}$. $\mathcal{G}_{n}$, $\mathcal{D}_{n}$ denotes the generative and the discriminator neural networks. The game minimizes the variance of the Earth-Mover distance(Wasserstein-1 metric, deduced from the Kantorovich-Rubinstein duality) between two multidimensional datasets(original \& imputed) based on probability distributions through the minimax(bi-level) optimization schema described by, 

\vspace{-3mm}
\begin{equation}
\min _{G_{n}} \max _{D_{n}}\big(\mathbb{E}_{\mathbf{x}^{t}_{i} \sim p(\mathbf{x}^{t}_{i})} \big[\mathbf{x}^{t}_{i}]-\mathbf{E}_{\mathbf{z}^{t}_{i} \sim {p(\mathbf{z}^{t}_{i})}} [\mathcal{D}_{n}(\mathcal{G}_{n}(\mathbf{z}^{t}_{i}))\big]\big) \label{eqn:gan}.
\end{equation}

\vspace{-2mm}
The loss function for $\mathcal{D}_{n}$ is described below, 

\vspace{-4mm}
\begin{equation}
\mathbb{E}_{\mathbf{x}^{t}_{i} \sim p(\mathbf{x}^{t}_{i})} \big[\mathbf{x}^{t}_{i}]-\mathbb{E}_{\mathbf{z}^{t}_{i} \sim {p(\mathbf{z}^{t}_{i})}} [\mathcal{D}_{n}(\mathcal{G}_{n}(\mathbf{z}^{t}_{i}))\big]
\end{equation}

\vspace{-1mm}
The loss function for $\mathcal{G}_{n}$ as follows,

\vspace{-4mm}
\begin{equation}
-\mathbb{E}_{\mathbf{z}^{t}_{i} \sim {p(\mathbf{z}^{t}_{i})}} [\mathcal{D}_{n}(\mathcal{G}_{n}(\mathbf{z}^{t}_{i}))\big]
\end{equation}

\vspace{-2mm}
$\mathbf{z}^{t}_{i} \in \mathbb{R}^{\mathcal{T} \cdot \mathcal{C}}$ is sampled from a uniform distribution. We model the $\small \mathcal{G}_{n}$ and $\small \mathcal{D}_{n}$ of the WGAN framework with the BGN model with different trainable parameters.  The input to the $\small \mathcal{G}_{n}$ is described below,

\vspace{-4mm}
\begin{equation}
\hat{\mathbf{z}}^{t}_{i} = \mathbf{m}^{t}_{i} \odot \mathbf{x}^{t}_{i} + (\mathbf{1}-\mathbf{m}^{t}_{i}) \odot \mathbf{z}^{t}_{i}
\end{equation}

\vspace{-2mm}
 $\mathbf{m}^{t}_{i} \in \{0,1\}$ denotes the random mask variable. The DGI module of $\small \mathcal{G}_{n}$ computes the sparse adjacency matrix $\hat{\mathcal{A}}^{(t)}$.  The grapher module determines the node-level representations $\hat{\mathbf{h}}_i^{(t)}$. We learn the node embeddings $\hat{\mathbf{b}}_i$ along with the other training parameters of the GAN. We compute the element-wise product of $\hat{\mathbf{h}}_i^{(t)}$ and $\hat{\mathbf{b}}_i$ to obtain the imputed data $\hat{\mathbf{x}}^{(t)}_i$. The imputed data is obtained by,

\vspace{-6mm}
\begin{align*}
\hat{\mathbf{x}}^{t}_{i} = \mathcal{G}_{n}(\mathbf{x}^{t}_{i}, \mathbf{m}^{t}_{i},(\mathbf{1}-\mathbf{m}^{t}_{i}) \odot \mathbf{z}^{t}_{i}) \\
\hat{\mathbf{x}}^{t}_{i} = \mathbf{m}^{t}_{i} \odot \mathbf{x}^{t}_{i} + (\mathbf{1}-\mathbf{m}^{t}_{i}) \odot \hat{\mathbf{x}}^{t}_{i}
\end{align*}

\vspace{-2mm}
The discriminator net, $\mathcal{D}_{n}: \hat{\mathbf{x}}^{t}_{i}  \rightarrow \hat{\mathbf{m}}^{t}_{i}$ takes as input the imputed data $\hat{\mathbf{x}}^{t}_{i}$ and outputs an estimated mask matrix $\hat{\mathbf{m}}^{t}_{i}$. The cross-entropy loss in binary classification is described by,

\vspace{-6mm}
\begin{align*}
-(\mathbf{m}^{t}_{i} \log \hat{\mathbf{m}}^{t}_{i} + (1-\mathbf{m}^{t}_{i}) \log (1-\hat{\mathbf{m}}^{t}_{i})) \mid \mathbf{m}^{t}_{i}=0
\end{align*}

\vspace{-2mm}
The generative neural network($\mathcal{G}_{n}$) minimizes the following loss function,

\vspace{-6mm}
\begin{align*}
\mathbf{m}^{t}_{i}(\mathbf{x}^{t}_{i} - \hat{\mathbf{x}}^{t}_{i})^{2}
\end{align*}

\vspace{-3mm}
It minimizes the imputation loss. We also minimize the Wasserstein distance between the estimates of two probability distributions $\small \hat{p}(\hat{\mathbf{x}}^{t}_{i})$ and $\small p(\mathbf{x}^{t}_{i})$. In summary, $\mathcal{G}_{n}$ and $\mathcal{D}_{n}$ are trained adversarially by deceptive input to impute the missing data.

\vspace{-3mm}
\begin{minipage}{0.56\textwidth}
\begin{algorithm}[H]
    \centering
    \caption{Generator network, $\small \mathcal{G}_{n}$}\label{algorithm}
    \begin{algorithmic}[1]
        \State \textbf{Input}  $\mathbf{x}^{t}_{i}, \mathbf{m}^{t}_{i}, \mathbf{z}^{t}_{i}$
        \State DGI, i.e., $\mathbf{m}^{t}_{i} \odot \mathbf{x}^{t}_{i} + (\mathbf{1}-\mathbf{m}^{t}_{i}) \odot \mathbf{z}^{t}_{i} \rightarrow  \hat{\mathbf{z}}^{t}_{i} \rightarrow \hat{\mathcal{A}}^{(t)}$
        \State Grapher, i.e., $\hat{\mathbf{z}}^{t}_{i},  \hat{\mathcal{A}}^{(t)}\rightarrow \hat{\mathbf{h}}_i^{(t)}$
        \State $\small \hat{\mathbf{x}}^{t}_{i} = \hat{\mathbf{h}}_i^{(t)} \odot \hat{\mathbf{b}}_i,  \mid i \in \mathcal{V}$
        \State \textbf{Return}  $\hat{\mathbf{x}}^{t}_{i} = \mathbf{m}^{t}_{i} \odot \mathbf{x}^{t}_{i} + (\mathbf{1}-\mathbf{m}^{t}_{i}) \odot \hat{\mathbf{x}}^{t}_{i}$
    \end{algorithmic}
\end{algorithm}
\end{minipage}
\hfill
\begin{minipage}{0.4\textwidth}
\begin{algorithm}[H]
    \centering
    \caption{Discriminator network, $\small \mathcal{D}_{n}$}\label{algorithm1}
    \begin{algorithmic}[1]
        \State \textbf{Input}  $\small \hat{\mathbf{x}}^{t}_{i}$
        \State Multi-layer Perceptron 
        \State non-linear sigmoid activation
        \State \textbf{Return}  $\hat{\mathbf{m}}^{t}_{i}$
    \end{algorithmic}
\end{algorithm}
\end{minipage}

\vspace{1mm}
\textbf{RUL Prediction :} We demonstrate the benefits of imputed data. We perform experiments on the NASA Randomized Battery Usage Dataset\cite{saha2007battery} and the UNIBO Powertools Dataset\cite{wong2021li}. The WGAN algorithm is trained  by randomly masking the training dataset to impute the masked data. We utilize the validation dataset for hyperparameter tuning. We obtain the imputed dataset corresponding to the training dataset from the WGAN framework. We select the BGN framework as a baseline model to learn from the training dataset for RUL prediction. The BGN$^*$ denotes the variant of the baseline model trained with the imputed  dataset for RUL prediction. We evaluate the performance of both models on the holdout dataset. \autoref{tab:imputeddata} reports the model's performance. We observe an on-par performance of the BGN$^*$ model w.r.t baseline framework on the NASA Randomized Battery\cite{saha2007battery} and UNIBO \cite{wong2021li} test datasets, respectively. The results show the effectiveness of our framework for imputing missing data.

\vspace{-1mm}
\begin{table}[htbp]
\centering
\resizebox{0.6\textwidth}{!}{%
\begin{tabular}{|c|c|c|c|c|c|c|c|c|}
\hline
Algorithms & $\textbf{BGN}$ & $\textbf{BGN}^*$    \\ \hline
 NASA Data Set\cite{saha2007battery}& \textbf{8.657    $\pm$  0.045} & \textbf{9.233   $\pm$  0.175}    \\ \hline
UNIBO Powertools Dataset\cite{wong2021li} &  \textbf{7.570  $\pm$  0.032}   & \textbf{8.806    $\pm$  0.118} \\ \hline
\end{tabular}%
}
\vspace{1mm}
\caption{\label{tab:imputeddata}Performance comparison for RUL prediction.}
\end{table}

\vspace{-5mm}
{
\small
\bibliographystyle{plain}
\bibliography{main.bib}

\begin{thebibliography}{10}

\bibitem{andre2013comparative}
D~Andre, A~Nuhic, T~Soczka-Guth, and Dirk~Uwe Sauer.
\newblock Comparative study of a structured neural network and an extended kalman filter for state of health determination of lithium-ion batteries in hybrid electricvehicles.
\newblock {\em Engineering Applications of Artificial Intelligence}, 26(3):951--961, 2013.

\bibitem{arjovsky2017wasserstein}
Martin Arjovsky, Soumith Chintala, and L{\'e}on Bottou.
\newblock Wasserstein generative adversarial networks.
\newblock In {\em International conference on machine learning}, pages 214--223. PMLR, 2017.

\bibitem{bai2020adaptive}
Lei Bai, Lina Yao, Can Li, Xianzhi Wang, and Can Wang.
\newblock Adaptive graph convolutional recurrent network for traffic forecasting.
\newblock {\em Advances in neural information processing systems}, 33:17804--17815, 2020.

\bibitem{baldacchino2016variational}
Tara Baldacchino, Elizabeth~J Cross, Keith Worden, and Jennifer Rowson.
\newblock Variational bayesian mixture of experts models and sensitivity analysis for nonlinear dynamical systems.
\newblock {\em Mechanical Systems and Signal Processing}, 66:178--200, 2016.

\bibitem{baldi2013understanding}
Pierre Baldi and Peter~J Sadowski.
\newblock Understanding dropout.
\newblock {\em Advances in neural information processing systems}, 26, 2013.

\bibitem{bengio2013estimating}
Yoshua Bengio, Nicholas L{\'e}onard, and Aaron Courville.
\newblock Estimating or propagating gradients through stochastic neurons for conditional computation.
\newblock {\em arXiv preprint arXiv:1308.3432}, 2013.

\bibitem{bianchi2021graph}
Filippo~Maria Bianchi, Daniele Grattarola, Lorenzo Livi, and Cesare Alippi.
\newblock Graph neural networks with convolutional arma filters.
\newblock {\em IEEE transactions on pattern analysis and machine intelligence}, 2021.

\bibitem{bielak2021graph}
Piotr Bielak, Tomasz Kajdanowicz, and Nitesh~V Chawla.
\newblock Graph barlow twins: A self-supervised representation learning framework for graphs.
\newblock {\em arXiv preprint arXiv:2106.02466}, 2021.

\bibitem{bosello2021charge}
Michael Bosello, Carlo Falcomer, Claudio Rossi, and Giovanni Pau.
\newblock To charge or to sell? ev pack useful life estimation via lstms and autoencoders.
\newblock {\em arXiv preprint arXiv:2110.03585}, 2021.

\bibitem{bresson2017residual}
Xavier Bresson and Thomas Laurent.
\newblock Residual gated graph convnets.
\newblock {\em arXiv preprint arXiv:1711.07553}, 2017.

\bibitem{chen2018gc}
Jinyin Chen, Xueke Wang, and Xuanheng Xu.
\newblock Gc-lstm: Graph convolution embedded lstm for dynamic link prediction.
\newblock {\em arXiv preprint arXiv:1812.04206}, 2018.

\bibitem{chen}
Ming Chen, Zhewei Wei, Zengfeng Huang, Bolin Ding, and Yaliang Li.
\newblock Simple and deep graph convolutional networks.
\newblock In {\em International Conference on Machine Learning}, pages 1725--1735. PMLR, 2020.

\bibitem{chen2019review}
Weidong Chen, Jun Liang, Zhaohua Yang, and Gen Li.
\newblock A review of lithium-ion battery for electric vehicle applications and beyond.
\newblock {\em Energy Procedia}, 158:4363--4368, 2019.

\bibitem{chen2020iterative}
Yu~Chen, Lingfei Wu, and Mohammed Zaki.
\newblock Iterative deep graph learning for graph neural networks: Better and robust node embeddings.
\newblock {\em Advances in neural information processing systems}, 33:19314--19326, 2020.

\bibitem{cho2014properties}
Kyunghyun Cho, Bart Van~Merri{\"e}nboer, Dzmitry Bahdanau, and Yoshua Bengio.
\newblock On the properties of neural machine translation: Encoder-decoder approaches.
\newblock {\em arXiv preprint arXiv:1409.1259}, 2014.

\bibitem{cho2014learning}
Kyunghyun Cho, Bart Van~Merri{\"e}nboer, Caglar Gulcehre, Dzmitry Bahdanau, Fethi Bougares, Holger Schwenk, and Yoshua Bengio.
\newblock Learning phrase representations using rnn encoder-decoder for statistical machine translation.
\newblock {\em arXiv preprint arXiv:1406.1078}, 2014.

\bibitem{chung2014empirical}
Junyoung Chung, Caglar Gulcehre, KyungHyun Cho, and Yoshua Bengio.
\newblock Empirical evaluation of gated recurrent neural networks on sequence modeling.
\newblock {\em arXiv preprint arXiv:1412.3555}, 2014.

\bibitem{cortes1995support}
Corinna Cortes and Vladimir Vapnik.
\newblock Support-vector networks.
\newblock {\em Machine learning}, 20(3):273--297, 1995.

\bibitem{daigle2016end}
Matthew Daigle and Chetan~S Kulkarni.
\newblock End-of-discharge and end-of-life prediction in lithium-ion batteries with electrochemistry-based aging models.
\newblock In {\em AIAA Infotech@ aerospace}, page 2132. 2016.

\bibitem{defferrard2016convolutional}
Micha{\"e}l Defferrard, Xavier Bresson, and Pierre Vandergheynst.
\newblock Convolutional neural networks on graphs with fast localized spectral filtering.
\newblock {\em Advances in neural information processing systems}, 29, 2016.

\bibitem{du2017topology}
Jian Du, Shanghang Zhang, Guanhang Wu, Jos{\'e}~MF Moura, and Soummya Kar.
\newblock Topology adaptive graph convolutional networks.
\newblock {\em arXiv preprint arXiv:1710.10370}, 2017.

\bibitem{el2020exploits}
Abdel El~Kharbachi, Olena Zavorotynska, M~Latroche, Ferm{\`\i}n Cuevas, Volodymyr Yartys, and M~Fichtner.
\newblock Exploits, advances and challenges benefiting beyond li-ion battery technologies.
\newblock {\em Journal of Alloys and Compounds}, 817:153261, 2020.

\bibitem{michael2022battery}
Michael~Bosello et~al.
\newblock Battery-state-estimation.
\newblock {\em GitHub. Note: https://github.com/MichaelBosello/battery-rul-estimation}, 2022.

\bibitem{fey2019just}
Matthias Fey.
\newblock Just jump: Dynamic neighborhood aggregation in graph neural networks.
\newblock {\em arXiv preprint arXiv:1904.04849}, 2019.

\bibitem{Fey/Lenssen/2019}
Matthias Fey and Jan~E. Lenssen.
\newblock Fast graph representation learning with {PyTorch Geometric}.
\newblock In {\em ICLR Workshop on Representation Learning on Graphs and Manifolds}, 2019.

\bibitem{franceschi2019learning}
Luca Franceschi, Mathias Niepert, Massimiliano Pontil, and Xiao He.
\newblock Learning discrete structures for graph neural networks.
\newblock In {\em International conference on machine learning}, pages 1972--1982. PMLR, 2019.

\bibitem{friedman2001greedy}
Jerome~H Friedman.
\newblock Greedy function approximation: a gradient boosting machine.
\newblock {\em Annals of statistics}, pages 1189--1232, 2001.

\bibitem{gao2019graph}
Hongyang Gao and Shuiwang Ji.
\newblock Graph u-nets.
\newblock In {\em international conference on machine learning}, pages 2083--2092. PMLR, 2019.

\bibitem{guttenberg2016permutation}
Nicholas Guttenberg, Nathaniel Virgo, Olaf Witkowski, Hidetoshi Aoki, and Ryota Kanai.
\newblock Permutation-equivariant neural networks applied to dynamics prediction.
\newblock {\em arXiv preprint arXiv:1612.04530}, 2016.

\bibitem{hasib2021comprehensive}
Shahid~A Hasib, S~Islam, Ripon~K Chakrabortty, Michael~J Ryan, Dip~Kumar Saha, Md~H Ahamed, Sumaya~I Moyeen, Sajal~K Das, Md~F Ali, Md~R Islam, et~al.
\newblock A comprehensive review of available battery datasets, rul prediction approaches, and advanced battery management.
\newblock {\em Ieee Access}, 2021.

\bibitem{hassani2020contrastive}
Kaveh Hassani and Amir~Hosein Khasahmadi.
\newblock Contrastive multi-view representation learning on graphs.
\newblock In {\em International Conference on Machine Learning}, pages 4116--4126. PMLR, 2020.

\bibitem{haykin1994neural}
Simon Haykin.
\newblock {\em Neural networks: a comprehensive foundation}.
\newblock Prentice Hall PTR, 1994.

\bibitem{hochreiter1997long}
Sepp Hochreiter and J{\"u}rgen Schmidhuber.
\newblock Long short-term memory.
\newblock {\em Neural computation}, 9(8):1735--1780, 1997.

\bibitem{hoshen2017vain}
Yedid Hoshen.
\newblock Vain: Attentional multi-agent predictive modeling.
\newblock {\em Advances in Neural Information Processing Systems}, 30, 2017.

\bibitem{ioffe2015batch}
Sergey Ioffe and Christian Szegedy.
\newblock Batch normalization: Accelerating deep network training by reducing internal covariate shift.
\newblock In {\em International conference on machine learning}, pages 448--456. PMLR, 2015.

\bibitem{jang2016categorical}
Eric Jang, Shixiang Gu, and Ben Poole.
\newblock Categorical reparameterization with gumbel-softmax.
\newblock {\em arXiv preprint arXiv:1611.01144}, 2016.

\bibitem{jiang2019semi}
Bo~Jiang, Ziyan Zhang, Doudou Lin, Jin Tang, and Bin Luo.
\newblock Semi-supervised learning with graph learning-convolutional networks.
\newblock In {\em Proceedings of the IEEE/CVF conference on computer vision and pattern recognition}, pages 11313--11320, 2019.

\bibitem{kazi2022differentiable}
Anees Kazi, Luca Cosmo, Seyed-Ahmad Ahmadi, Nassir Navab, and Michael Bronstein.
\newblock Differentiable graph module (dgm) for graph convolutional networks.
\newblock {\em IEEE Transactions on Pattern Analysis and Machine Intelligence}, 2022.

\bibitem{kim2022find}
Dongkwan Kim and Alice Oh.
\newblock How to find your friendly neighborhood: Graph attention design with self-supervision.
\newblock {\em arXiv preprint arXiv:2204.04879}, 2022.

\bibitem{kim2021atom}
Se-Ho Kim, Stoichko Antonov, Xuyang Zhou, Leigh~T Stephenson, Chanwon Jung, Ayman~A El-Zoka, Daniel~K Schreiber, Michele Conroy, and Baptiste Gault.
\newblock Atom probe analysis of battery materials: challenges and ways forward.
\newblock {\em arXiv preprint arXiv:2110.03716}, 2021.

\bibitem{kingma2014adam}
Diederik~P Kingma and Jimmy Ba.
\newblock Adam: A method for stochastic optimization.
\newblock {\em arXiv preprint arXiv:1412.6980}, 2014.

\bibitem{kingma2013auto}
Diederik~P Kingma and Max Welling.
\newblock Auto-encoding variational bayes.
\newblock {\em arXiv preprint arXiv:1312.6114}, 2013.

\bibitem{kipf2016semi}
Thomas~N Kipf and Max Welling.
\newblock Semi-supervised classification with graph convolutional networks.
\newblock {\em arXiv preprint arXiv:1609.02907}, 2016.

\bibitem{klicpera2018predict}
Johannes Klicpera, Aleksandar Bojchevski, and Stephan G{\"u}nnemann.
\newblock Predict then propagate: Graph neural networks meet personalized pagerank.
\newblock {\em arXiv preprint arXiv:1810.05997}, 2018.

\bibitem{kool2019stochastic}
Wouter Kool, Herke Van~Hoof, and Max Welling.
\newblock Stochastic beams and where to find them: The gumbel-top-k trick for sampling sequences without replacement.
\newblock In {\em International Conference on Machine Learning}, pages 3499--3508. PMLR, 2019.

\bibitem{li2019predicting}
Jia Li, Zhichao Han, Hong Cheng, Jiao Su, Pengyun Wang, Jianfeng Zhang, and Lujia Pan.
\newblock Predicting path failure in time-evolving graphs.
\newblock In {\em Proceedings of the 25th ACM SIGKDD International Conference on Knowledge Discovery \& Data Mining}, pages 1279--1289, 2019.

\bibitem{li2017diffusion}
Yaguang Li, Rose Yu, Cyrus Shahabi, and Yan Liu.
\newblock Diffusion convolutional recurrent neural network: Data-driven traffic forecasting.
\newblock {\em arXiv preprint arXiv:1707.01926}, 2017.

\bibitem{li2015gated}
Yujia Li, Daniel Tarlow, Marc Brockschmidt, and Richard Zemel.
\newblock Gated graph sequence neural networks.
\newblock {\em arXiv preprint arXiv:1511.05493}, 2015.

\bibitem{liaw2002classification}
Andy Liaw, Matthew Wiener, et~al.
\newblock Classification and regression by randomforest.
\newblock {\em R news}, 2(3):18--22, 2002.

\bibitem{liu2019brief}
Kailong Liu, Kang Li, Qiao Peng, and Cheng Zhang.
\newblock A brief review on key technologies in the battery management system of electric vehicles.
\newblock {\em Frontiers of mechanical engineering}, 14(1):47--64, 2019.

\bibitem{maddison2016concrete}
Chris~J Maddison, Andriy Mnih, and Yee~Whye Teh.
\newblock The concrete distribution: A continuous relaxation of discrete random variables.
\newblock {\em arXiv preprint arXiv:1611.00712}, 2016.

\bibitem{manthiram2017outlook}
Arumugam Manthiram.
\newblock An outlook on lithium ion battery technology.
\newblock {\em ACS central science}, 3(10):1063--1069, 2017.

\bibitem{mohtat2021algorithmic}
Peyman Mohtat, Sravan Pannala, Valentin Sulzer, Jason~B Siegel, and Anna~G Stefanopoulou.
\newblock An algorithmic safety vest for li-ion batteries during fast charging.
\newblock {\em IFAC-PapersOnLine}, 54(20):522--527, 2021.

\bibitem{morris2019weisfeiler}
Christopher Morris, Martin Ritzert, Matthias Fey, William~L Hamilton, Jan~Eric Lenssen, Gaurav Rattan, and Martin Grohe.
\newblock Weisfeiler and leman go neural: Higher-order graph neural networks.
\newblock In {\em Proceedings of the AAAI conference on artificial intelligence}, volume~33, pages 4602--4609, 2019.

\bibitem{nix1994estimating}
David~A Nix and Andreas~S Weigend.
\newblock Estimating the mean and variance of the target probability distribution.
\newblock In {\em Proceedings of 1994 ieee international conference on neural networks (ICNN'94)}, volume~1, pages 55--60. IEEE, 1994.

\bibitem{nuhic2018battery}
Adnan Nuhic, Jonas Bergdolt, Bernd Spier, Michael Buchholz, and Klaus Dietmayer.
\newblock Battery health monitoring and degradation prognosis in fleet management systems.
\newblock {\em World Electric Vehicle Journal}, 9(3):39, 2018.

\bibitem{o2022lithium}
Simon~EJ O'Kane, Weilong Ai, Ganesh Madabattula, Diego Alonso-Alvarez, Robert Timms, Valentin Sulzer, Jacqueline~Sophie Edge, Billy Wu, Gregory~J Offer, and Monica Marinescu.
\newblock Lithium-ion battery degradation: how to model it.
\newblock {\em Physical Chemistry Chemical Physics}, 24(13):7909--7922, 2022.

\bibitem{panagopoulos2021transfer}
George Panagopoulos, Giannis Nikolentzos, and Michalis Vazirgiannis.
\newblock Transfer graph neural networks for pandemic forecasting.
\newblock In {\em Proceedings of the AAAI Conference on Artificial Intelligence}, volume~35, pages 4838--4845, 2021.

\bibitem{peng2021conformer}
Zhiliang Peng, Wei Huang, Shanzhi Gu, Lingxi Xie, Yaowei Wang, Jianbin Jiao, and Qixiang Ye.
\newblock Conformer: Local features coupling global representations for visual recognition.
\newblock In {\em Proceedings of the IEEE/CVF International Conference on Computer Vision}, pages 367--376, 2021.

\bibitem{rozemberczki2021pytorch}
Benedek Rozemberczki, Paul Scherer, Yixuan He, George Panagopoulos, Alexander Riedel, Maria Astefanoaei, Oliver Kiss, Ferenc Beres, , Guzman Lopez, Nicolas Collignon, and Rik Sarkar.
\newblock {PyTorch Geometric Temporal: Spatiotemporal Signal Processing with Neural Machine Learning Models}.
\newblock In {\em Proceedings of the 30th ACM International Conference on Information and Knowledge Management}, page 4564–4573, 2021.

\bibitem{saha2007battery}
Bhaskar Saha and Kai Goebel.
\newblock Battery data set.
\newblock {\em NASA AMES prognostics data repository}, 2007.

\bibitem{saha2009modeling}
Bhaskar Saha and Kai Goebel.
\newblock Modeling li-ion battery capacity depletion in a particle filtering framework.
\newblock In {\em Annual Conference of the PHM Society}, volume~1, 2009.

\bibitem{santoro2017simple}
Adam Santoro, David Raposo, David~G Barrett, Mateusz Malinowski, Razvan Pascanu, Peter Battaglia, and Timothy Lillicrap.
\newblock A simple neural network module for relational reasoning.
\newblock {\em Advances in neural information processing systems}, 30, 2017.

\bibitem{schapire2013explaining}
Robert~E Schapire.
\newblock Explaining adaboost.
\newblock In {\em Empirical inference}, pages 37--52. Springer, 2013.

\bibitem{seo2018structured}
Youngjoo Seo, Micha{\"e}l Defferrard, Pierre Vandergheynst, and Xavier Bresson.
\newblock Structured sequence modeling with graph convolutional recurrent networks.
\newblock In {\em International conference on neural information processing}, pages 362--373. Springer, 2018.

\bibitem{shang2021discrete}
Chao Shang, Jie Chen, and Jinbo Bi.
\newblock Discrete graph structure learning for forecasting multiple time series.
\newblock {\em arXiv preprint arXiv:2101.06861}, 2021.

\bibitem{shen2019review}
Ming Shen and Qing Gao.
\newblock A review on battery management system from the modeling efforts to its multiapplication and integration.
\newblock {\em International Journal of Energy Research}, 43(10):5042--5075, 2019.

\bibitem{sukhbaatar2016learning}
Sainbayar Sukhbaatar, Rob Fergus, et~al.
\newblock Learning multiagent communication with backpropagation.
\newblock {\em Advances in neural information processing systems}, 29, 2016.

\bibitem{sun2019infograph}
Fan-Yun Sun, Jordan Hoffmann, Vikas Verma, and Jian Tang.
\newblock Infograph: Unsupervised and semi-supervised graph-level representation learning via mutual information maximization.
\newblock {\em arXiv preprint arXiv:1908.01000}, 2019.

\bibitem{taheri2019learning}
Aynaz Taheri, Kevin Gimpel, and Tanya Berger-Wolf.
\newblock Learning to represent the evolution of dynamic graphs with recurrent models.
\newblock In {\em Companion proceedings of the 2019 world wide web conference}, pages 301--307, 2019.

\bibitem{thakoor2021bootstrapped}
Shantanu Thakoor, Corentin Tallec, Mohammad~Gheshlaghi Azar, R{\'e}mi Munos, Petar Veli{\v{c}}kovi{\'c}, and Michal Valko.
\newblock Bootstrapped representation learning on graphs.
\newblock In {\em ICLR 2021 Workshop on Geometrical and Topological Representation Learning}, 2021.

\bibitem{thekumparampil2018attention}
Kiran~K Thekumparampil, Chong Wang, Sewoong Oh, and Li-Jia Li.
\newblock Attention-based graph neural network for semi-supervised learning.
\newblock {\em arXiv preprint arXiv:1803.03735}, 2018.

\bibitem{ulvestad2018brief}
Andrew Ulvestad.
\newblock A brief review of current lithium ion battery technology and potential solid state battery technologies.
\newblock {\em arXiv preprint arXiv:1803.04317}, 2018.

\bibitem{vaswani2017attention}
Ashish Vaswani, Noam Shazeer, Niki Parmar, Jakob Uszkoreit, Llion Jones, Aidan~N Gomez, {\L}ukasz Kaiser, and Illia Polosukhin.
\newblock Attention is all you need.
\newblock {\em Advances in neural information processing systems}, 30, 2017.

\bibitem{velivckovic2017graph}
Petar Veli{\v{c}}kovi{\'c}, Guillem Cucurull, Arantxa Casanova, Adriana Romero, Pietro Lio, and Yoshua Bengio.
\newblock Graph attention networks.
\newblock {\em arXiv preprint arXiv:1710.10903}, 2017.

\bibitem{wang2020gcn}
Xiao Wang, Meiqi Zhu, Deyu Bo, Peng Cui, Chuan Shi, and Jian Pei.
\newblock Am-gcn: Adaptive multi-channel graph convolutional networks.
\newblock In {\em Proceedings of the 26th ACM SIGKDD International conference on knowledge discovery \& data mining}, pages 1243--1253, 2020.

\bibitem{watters2017visual}
Nicholas Watters, Daniel Zoran, Theophane Weber, Peter Battaglia, Razvan Pascanu, and Andrea Tacchetti.
\newblock Visual interaction networks: Learning a physics simulator from video.
\newblock {\em Advances in neural information processing systems}, 30, 2017.

\bibitem{wong2021li}
Kei~Long Wong, Michael Bosello, Rita Tse, Carlo Falcomer, Claudio Rossi, and Giovanni Pau.
\newblock Li-ion batteries state-of-charge estimation using deep lstm at various battery specifications and discharge cycles.
\newblock In {\em Proceedings of the Conference on Information Technology for Social Good}, pages 85--90, 2021.

\bibitem{woo2022etsformer}
Gerald Woo, Chenghao Liu, Doyen Sahoo, Akshat Kumar, and Steven Hoi.
\newblock Etsformer: Exponential smoothing transformers for time-series forecasting.
\newblock {\em arXiv preprint arXiv:2202.01381}, 2022.

\bibitem{wu2016online}
Ji~Wu, Chenbin Zhang, and Zonghai Chen.
\newblock An online method for lithium-ion battery remaining useful life estimation using importance sampling and neural networks.
\newblock {\em Applied energy}, 173:134--140, 2016.

\bibitem{xiong2018towards}
Rui Xiong, Linlin Li, and Jinpeng Tian.
\newblock Towards a smarter battery management system: A critical review on battery state of health monitoring methods.
\newblock {\em Journal of Power Sources}, 405:18--29, 2018.

\bibitem{you2020graph}
Yuning You, Tianlong Chen, Yongduo Sui, Ting Chen, Zhangyang Wang, and Yang Shen.
\newblock Graph contrastive learning with augmentations.
\newblock {\em Advances in Neural Information Processing Systems}, 33:5812--5823, 2020.

\bibitem{zhang2020gnnguard}
Xiang Zhang and Marinka Zitnik.
\newblock Gnnguard: Defending graph neural networks against adversarial attacks.
\newblock {\em Advances in neural information processing systems}, 33:9263--9275, 2020.

\bibitem{zhao2019t}
Ling Zhao, Yujiao Song, Chao Zhang, Yu~Liu, Pu~Wang, Tao Lin, Min Deng, and Haifeng Li.
\newblock T-gcn: A temporal graph convolutional network for traffic prediction.
\newblock {\em IEEE Transactions on Intelligent Transportation Systems}, 21(9):3848--3858, 2019.

\bibitem{zheng2020robust}
Cheng Zheng, Bo~Zong, Wei Cheng, Dongjin Song, Jingchao Ni, Wenchao Yu, Haifeng Chen, and Wei Wang.
\newblock Robust graph representation learning via neural sparsification.
\newblock In {\em International Conference on Machine Learning}, pages 11458--11468. PMLR, 2020.

\bibitem{zhou2021lithium}
Kate~Qi Zhou, Yan Qin, Billy Pik~Lik Lau, Chau Yuen, and Stefan Adams.
\newblock Lithium-ion battery state of health estimation based on cycle synchronization using dynamic time warping.
\newblock In {\em IECON 2021--47th Annual Conference of the IEEE Industrial Electronics Society}, pages 1--6. IEEE, 2021.

\bibitem{Zhu:2021tu}
Yanqiao Zhu, Yichen Xu, Qiang Liu, and Shu Wu.
\newblock {An Empirical Study of Graph Contrastive Learning}.
\newblock {\em arXiv.org}, September 2021.

\bibitem{zhu2020deep}
Yanqiao Zhu, Yichen Xu, Feng Yu, Qiang Liu, Shu Wu, and Liang Wang.
\newblock Deep graph contrastive representation learning.
\newblock {\em arXiv preprint arXiv:2006.04131}, 2020.

\bibitem{zubi2018lithium}
Ghassan Zubi, Rodolfo Dufo-L{\'o}pez, Monica Carvalho, and Guzay Pasaoglu.
\newblock The lithium-ion battery: State of the art and future perspectives.
\newblock {\em Renewable and Sustainable Energy Reviews}, 89:292--308, 2018.

\end{thebibliography}
}

\end{document}